\documentclass[10pt,twocolumn,letterpaper,dvipsnames,table]{article}

\usepackage[pagenumbers]{cvpr} % To force page numbers, e.g. for an arXiv version
\usepackage[accsupp]{axessibility}  % Improves PDF readability for those with disabilities.

\usepackage{cuted}
\usepackage{lineno}
\usepackage{pifont}
\usepackage{graphicx}
\usepackage{listings}
\usepackage{bbding}
\usepackage{algorithm}
\usepackage{algorithmic}
\usepackage{amsmath}
\usepackage{makecell}
\usepackage{amssymb}
\usepackage{booktabs}
\usepackage[percent]{overpic}  
\usepackage{multirow}
\usepackage{soul}
\usepackage[normalem]{ulem}

\definecolor{cvprblue}{rgb}{0.21,0.49,0.74}
\usepackage[pagebackref,breaklinks,colorlinks,allcolors=cvprblue]{hyperref}

\newcommand{\authorsep}{\hspace{8pt}}
\newcommand{\affiliationsep}{\hspace{8pt}}

\newcommand{\mygray}{\rowcolor[HTML]{f5f5f5}}
\newcommand{\bestcolor}[1]{\textbf{\color[HTML]{000000} #1}}
\newcommand{\secondcolor}[1]{\underline{\color[HTML]{000000} #1}}

\def\paperID{}
\def\confName{CVPR}
\def\confYear{2026}

\title{\parbox{0.8\linewidth}{
\centering
FlowDIS: Language-Guided Dichotomous Image Segmentation with Flow Matching
\vspace{-0.5cm}
}}

\author{Andranik Sargsyan \authorsep
Shant Navasardyan \\
\small Picsart AI Research (PAIR) \affiliationsep
\\{\small \textbf{\url{https://flowdis.github.io}}}
}

\begin{document}
\maketitle

% Teaser figure
\begin{strip}
\vspace{-15mm}
    \centering
    \includegraphics[width=\linewidth]{main/figures/teaser.pdf}
    \captionof{figure}{\textbf{\textit{FlowDIS}} enables highly accurate foreground segmentation, \textit{optionally} guided by a text prompt (the first example does not use text prompt).
    When ambiguity prevents the model from producing the desired result, the user can specify which elements to retain in the foreground. 
    For example, in the upper-right image, if the user wants the bicycle to be segmented, they can provide the prompt “bicycle” to FlowDIS and obtain a binary mask containing only the pixels shown in cyan. 
    Alternatively, if the user wants all three main objects to be segmented, they can prompt with “bicycle, tree, and white ladder”, and FlowDIS will return a combined binary mask including all the pixels shown in cyan, green, and red.
    }
    \label{fig:teaser-image}
\end{strip}

\begin{abstract}
Accurate image segmentation is essential for modern computer vision applications such as image editing, autonomous driving, and medical image analysis. In recent years, Dichotomous Image Segmentation (DIS)  has become a standard task for training and evaluating highly accurate segmentation models. 
Existing DIS approaches often fail to preserve fine-grained details or fully capture the semantic structure of the foreground.
To address these challenges, we present \textbf{FlowDIS}, a novel dichotomous image segmentation method built on the flow matching framework, which learns a time-dependent vector field to transport the image distribution to the corresponding mask distribution, optionally conditioned on a text prompt.
Moreover, with our \textbf{Position-Aware Instance Pairing (PAIP)} training strategy, FlowDIS offers strong controllability through text prompts, enabling precise, pixel-level object segmentation.
Extensive experiments demonstrate that our method significantly outperforms state-of-the-art approaches both with and without language guidance. Compared with the best prior DIS method, FlowDIS achieves a $\textbf{5.5\%}$ \textbf{higher $F_\beta^\omega$} measure and $\textbf{43\%}$ \textbf{lower MAE ($\mathcal{M}$)} on the DIS-TE test set.
The code is available at:  \href{https://github.com/Picsart-AI-Research/FlowDIS}{https://github.com/Picsart-AI-Research/FlowDIS}.
\end{abstract}
    
\section{Introduction}
\label{sec:intro}

Highly accurate image segmentation is crucial for modern computer vision applications, where even small errors can significantly impact downstream tasks such as image editing \cite{liu2025step1x, lu2025pinco}, autonomous driving \cite{li2025weakly}, and medical image analysis \cite{ji2024frontiers}. Dichotomous Image Segmentation (DIS) \cite{qin2022highly}, which involves segmenting high-precision, category-agnostic masks, has become an increasingly popular direction in the research community for developing and evaluating high-accuracy segmentation models. 

Many DIS methods \cite{qin2022highly, kim2022revisiting, zhou2023dichotomous, zheng2024birefnet, yu2024multi} treat segmentation as a per-pixel binary classification task and rely on pre-trained classification networks such as ResNet \cite{he2016deep}, Res2Net \cite{gao2019res2net}, or Swin Transformer \cite{liu2021swin} as backbones. 
However, since these backbones are optimized for predicting the overall class of an image, they often lack the fine-grained semantic representations required for accurate foreground segmentation, resulting in suboptimal performance on images with intricate details.
Moreover, in real-world scenes containing multiple objects, classification backbones often struggle to identify and segment the correct foreground regions due to their limited ability to capture object-level semantics.

Recent progress in generative modeling \cite{rombach2022high, podell2023sdxl} has motivated the formulation of image segmentation within the DDPM framework \cite{ho2020denoising}, allowing DIS models to leverage pre-trained text-to-image (T2I) diffusion priors rather than conventional classification backbones. T2I generative models, trained on large-scale and semantically diverse datasets, provide rich representations that have proven beneficial for the downstream segmentation task.
In particular, DiffDIS \cite{DiffDIS} and LawDIS \cite{yan2025lawdis} frame image segmentation as image-conditioned mask generation from standard Gaussian noise, building upon pre-trained Stable Diffusion \cite{rombach2022high}.

However, as also noted by prior work \cite{pang2025aligning111,xia2025mathrm111,lee2024exploiting111}, stochastic generative formulations are misaligned with deterministic dense prediction tasks, such as image segmentation, that require precise matches to the ground truth. This discrepancy often results in slower training convergence, requiring tens of thousands of optimization steps. Furthermore, the stochastic nature of the denoising process can blur or misplace fine boundaries, degrading the segmentation of intricate foreground structures.

To address these issues, we observe that image segmentation can be more naturally framed within the flow matching framework, enabling fully deterministic training and sampling. 
Flow matching offers a general framework for learning mappings between arbitrary distributions.
Based on this, we propose \textbf{\textit{FlowDIS}}, which directly learns the mapping from the image distribution to the corresponding mask distribution, in contrast to the image-conditioned mask denoising (generation) paradigm of existing diffusion-based DIS methods.

Another notable property of generative models such as diffusion or flow matching models is their strong adaptability to text prompts.
When adapted for the DIS task, this capability can be leveraged, as demonstrated by \cite{yan2025lawdis}, to mitigate the ambiguity caused by multiple foreground objects within a single image.
However, due to the complex multi-object nature of real-world scenes and the limited presence of multi-foreground examples in standard DIS training datasets, straightforward prompt-guided training still fails to achieve reliable language controllability, even with extensive prompting. 
Therefore, to further improve the language controllability, we propose the \textbf{\textit{Position-Aware Instance Pairing (PAIP)}} strategy, which constructs mixed training examples from pairs of \textit{(image, mask, prompt)} triplets within each training batch to provide the model with more diverse samples.
Fig. \ref{fig:teaser-image} demonstrates the capabilities of FlowDIS for different use cases, including background removal and language-guided segmentation.

\begin{figure*}[ht]
    \centering
    \includegraphics[width=\linewidth]{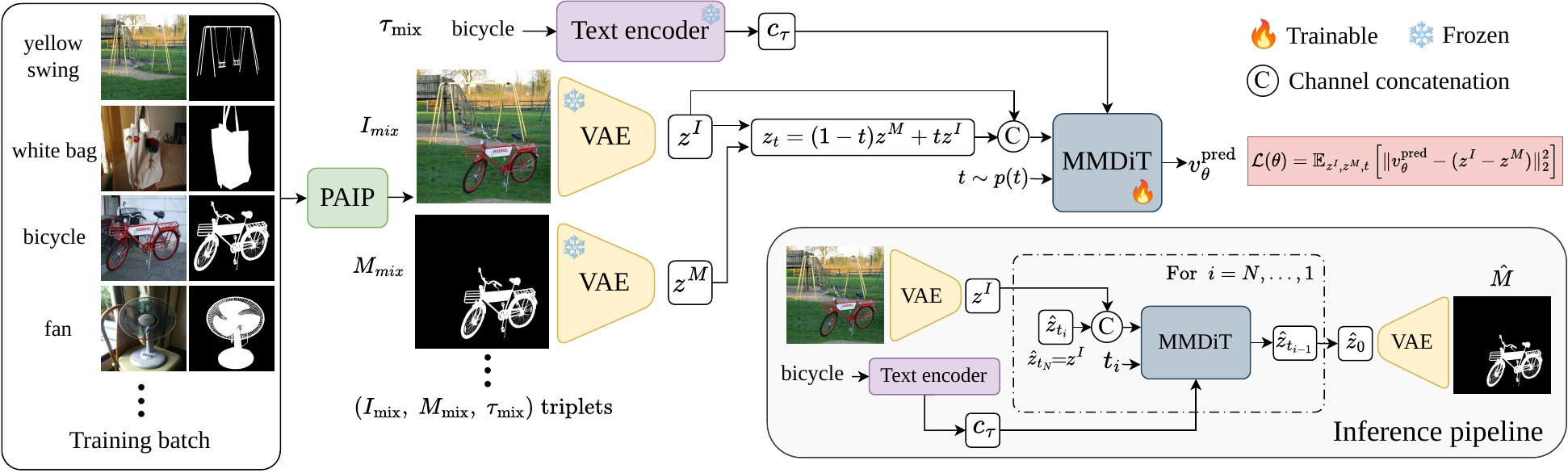}
    \caption{\textbf{Overall framework of FlowDIS.} During training, a batch of samples is passed to PAIP, which selectively combines pairs of \textit{(image, mask, prompt)} triplets to produce a mixed batch. The mixed images and masks are encoded into the VAE latent space. For timesteps $t \sim p(t)$, intermediate latents $z_{t}$ are obtained as a linear interpolation between the image and mask latents. Text prompts are encoded by the text encoder, and the resulting tokens $c_{\tau}$, together with $z_{t}$, $z^I$ and the sampled timesteps $t$, are fed into the MMDiT velocity prediction model. The training loss is computed as the MSE between the predicted and ground-truth velocities. During inference, the probability flow ODE is iteratively solved for $\hat{z}_0$ with the initial condition $\hat{z}_{1} = z^{I}$, where $z^{I}$ denotes the VAE encoding of the input image. The resulting latent $\hat{z}_0$ is then decoded by the VAE decoder to obtain the final mask prediction.
    }
    \label{figures:method_training}
\end{figure*}

To summarize, our main contributions are the following:
\begin{itemize}
\item We introduce \textbf{\textit{FlowDIS}}, a highly accurate dichotomous image segmentation model leveraging the power of flow-based generative modeling.
\item To enhance the language controllability of the model, we introduce a \textbf{\textit{Position-Aware Instance Pairing (PAIP)}}
strategy that selectively combines pairs of \textit{(image, mask, prompt)} triplets within each training batch.
\item FlowDIS establishes a new state-of-the-art across all test sets of DIS5K, outperforming previous methods by a significant margin. In particular, it achieves a $\textbf{5.5\%}$ higher $F_\beta^\omega$ score and a $\textbf{43\%}$ lower MAE ($\mathcal{M}$) on the DIS-TE test set compared to the best prior method.
\end{itemize}
\section{Related Work}
\label{sec:formatting}

Research in image segmentation has evolved into several specialized subfields, each focusing on distinct types of foreground detection—such as salient \cite{li2016deep, wang2019salient, zeng2019towards, tang2021disentangled}, camouflaged \cite{mei2021camouflaged, he2023camouflaged, chen2024camodiffusion}, and fine-grained \cite{liew2021deep, yang2020meticulous} object segmentation. To bridge these data-specific segmentation tasks, Qin et al. \cite{qin2022highly} introduced the Dichotomous Image Segmentation (DIS) task and the DIS5K dataset, which aim to provide a unified formulation and benchmark for highly accurate general foreground detection.

In recent years, DIS has attracted growing interest and motivated the development of several improved architectures. InSPyReNet \cite{kim2022revisiting} constructs a saliency map in an image‑pyramid structure, enabling the blending of low-resolution and high-resolution scale outputs via pyramid‑based image blending. FP-DIS \cite{zhou2023dichotomous} uses frequency priors to help the model capture more detailed information. Pei et al. \cite{pei2023unite} propose a dual-input network to disentangle the trunk and structure segmentation. BiRefNet \cite{zheng2024birefnet} incorporates a bilateral reference module that draws attention to detail-rich areas during training. MVANet \cite{yu2024multi} unifies distant-view and close-up feature fusion into a single encoder–decoder architecture. PDFNet \cite{liu2025highprecisiondichotomousimagesegmentation} introduces the depth-integrity prior to reduce false positive detections.

Recent advances in diffusion-based generative modeling \cite{ho2020denoising, rombach2022high} have motivated the development of diffusion-based DIS methods \cite{xu2024diffusion, DiffDIS, yan2025lawdis}. GenPercept \cite{xu2024diffusion} fine-tunes Stable Diffusion \cite{rombach2022high} for a number of dense prediction tasks, including DIS. DiffDIS \cite{DiffDIS} uses an auxiliary edge generation task and introduces a detail-balancing attention mechanism to improve the precision of the mask prediction. LawDIS \cite{yan2025lawdis} integrates user controls through language guidance and window refinement. Unlike diffusion-based DIS methods, our flow matching formulation offers a more intuitive framework for image segmentation and demonstrates superior empirical performance. 
\section{Method}

The overview of our method is shown in \cref{figures:method_training}, which builds on the flow matching \cite{Lipman2022FlowMF} framework. We first outline the general formulation of flow matching (\cref{sec:method:flow_matching}), then introduce our method FlowDIS in \cref{sec:method:flow_dis}. 
Finally, in \cref{sec:method:paip} we present a Position-Aware Instance Pairing (PAIP) strategy that enhances language controllability when training data are limited.

\subsection{Overview of Flow Matching}
\label{sec:method:flow_matching}

Flow matching \cite{Lipman2022FlowMF} is a generative modeling approach that learns a time-dependent vector field $v_\theta(x, t)$ whose trajectories transport samples from an easy-to-sample reference distribution $p_1(x)$ to a target distribution $p_0(x)$.

Since directly learning global trajectories is intractable, flow matching instead defines a \textit{conditional flow} between sample pairs $(x_0, x_1)$ and trains $v_\theta$ to match the induced velocity field along the resulting path $\{x_t\}_{t\in[0,1]}$.
A common choice for this conditional flow is the \textit{linear interpolation} between the samples $x_0 \sim p_0$ and $x_1 \sim p_1$:
\begin{equation}
x_t = (1 - t)x_0 + t x_1, \quad t \in [0, 1].
\end{equation}
The corresponding \textit{target velocity} along this path is
\begin{equation}
v(x_0, x_1) = \frac{d x_t}{d t} = x_1 - x_0.
\end{equation}
The model $v_\theta(x, t)$ is trained to approximate this conditional vector field in expectation over $p_0$, $p_1$, and $t$, by minimizing the \textit{flow matching loss}:
\begin{equation}
\label{eq:flow_loss}
\mathcal{L}(\theta) = \mathbb{E}_{\; x_0 \sim p_0, x_1 \sim p_1,t}
\Big[\, \| v_\theta(x_t, t) - v(x_0, x_1) \|^2_2 \,\Big]
\end{equation}
where $t\sim p(t)$ is the timestep sampling distribution during training.

Once trained, $v_\theta$ defines a continuous flow that can be integrated \textit{backward in time} from $t = 1$ to $t = 0$ to transform samples from $p_1$ into realistic samples from $p_0$:
\begin{equation}
\label{eq:flow_integral}
x_0 = x_1 + \int_{1}^{0} v_\theta(x_t, t)\, dt.
\end{equation}
In practice, this integration is performed numerically using a discretized solver such as the Euler method. Starting from a sample $x_1 \sim p_1$, the sampling process iteratively updates
\begin{equation}
x_{t - \Delta t} = x_t - \Delta t \, v_\theta(x_t, t),
\end{equation}
until $t = 0$, yielding a generated sample $x_0$ from the target distribution.

Notably, flow matching provides a unified formulation of generative modeling in which the reference distribution $p_1$ can be any chosen distribution, not limited to a Gaussian prior. Diffusion models can be viewed as a special case of flow matching, where the reference dynamics are stochastic and $p_1$ is a standard normal distribution \cite{Lipman2022FlowMF}.

\subsection{FlowDIS}
\label{sec:method:flow_dis}

FlowDIS leverages the power of flow matching by considering $p_1$ as the distribution of RGB images and $p_0$ as the distribution of the binary masks. Within this formulation, the velocity network is trained to transport an image $I \sim p_1$ toward its segmentation mask $M \sim p_0$, thereby learning to perform the segmentation task through deterministic flow-based generation. An overview of FlowDIS is illustrated in Fig.~\ref{figures:method_training}.

More precisely, given an input image $I \in \mathbb{R}^{H \times W \times 3}$ and its corresponding ground-truth mask $M \in \{0,1\}^{H \times W}$, both are encoded into a shared latent space using the encoder $\mathcal{E}(\cdot)$ of a variational autoencoder (VAE), producing the corresponding latent representations $z^I = \mathcal{E}(I)$ and $z^M = \mathcal{E}(M)$.

Following the standard optimal transport formulation of flow matching, we define the latent trajectory $z_t$ as:
\begin{equation}
z_t = (1 - t)z^{M} + tz^{I}, \quad t \in [0,1].
\end{equation}
In addition, to ensure that the velocity network always has access to the clean image signal, we condition it on the image latent $z^I$ by concatenating it to the input\footnote{An ablation study for this setting is included in the appendix.}, so the flow matching loss in \cref{eq:flow_loss} becomes:
\begin{equation}
\label{eq:flowdis_loss}
\begin{split}
    \mathcal{L}(\theta) = \mathbb{E}_{z^I, z^M, t}
    \left[\| v_\theta(z_t, z^I, t) - (z^{I} - z^{M}) \|_2^2\right]
\end{split}
\end{equation}

Moreover, FlowDIS extends this formulation with language-based conditioning, where the velocity network $v_\theta$ additionally takes the text embeddings $c_\tau$ of the corresponding text prompt $\tau$ as input, enabling language-guided segmentation and resulting in the final loss function:
\begin{equation}
\label{eq:flowdis_loss_lang}
\begin{split}
    \mathcal{L}(\theta) = \mathbb{E}_{z^I, z^M, t}
    \left[\| v_\theta(z_t, z^I, t, c_\tau) - (z^{I} - z^{M}) \|_2^2\right]
\end{split}
\end{equation}

\noindent
\textbf{Inference of FlowDIS:}
\noindent
Now let $I$ be an image with one or more objects, and $\tau$ be a text prompt indicating the object the user wants to segment from $I$. Then FlowDIS allows segmentation by solving the probability flow ODE:
\begin{equation}
    \label{eq:inference_ode}
    \frac{d z_t}{d t} = v_{\theta}(z_t, z^I, t, c_{\tau}), \quad z_1 = z^I,
\end{equation}
where $z^I=\mathcal{E}(I)$ is the latent representation of the inference image and $c_{\tau}$ is the encoding of the given language prompt $\tau$. 
To solve \cref{eq:inference_ode}, we use the Euler integration method, which requires a time discretization 
$0 = t_0 < t_1 < \ldots < t_N = 1$, yielding the following Euler integration step:
\begin{gather}
z_{t_i} = z_{t_{i+1}} + v_\theta(z_{t_{i+1}}, z^I, t_{i+1}, c_{\tau}) \,(t_i - t_{i+1})
\end{gather}
for $i=N-1, \ldots, 1,0$.
After completing all steps, the final latent $z_0$ is decoded with the VAE decoder $\mathcal{D}(\cdot)$ to produce the predicted mask.

\subsection{Position-Aware Instance Pairing}
\label{sec:method:paip}

\begin{figure*}[ht!]
    \centering
    \includegraphics[width=\linewidth]{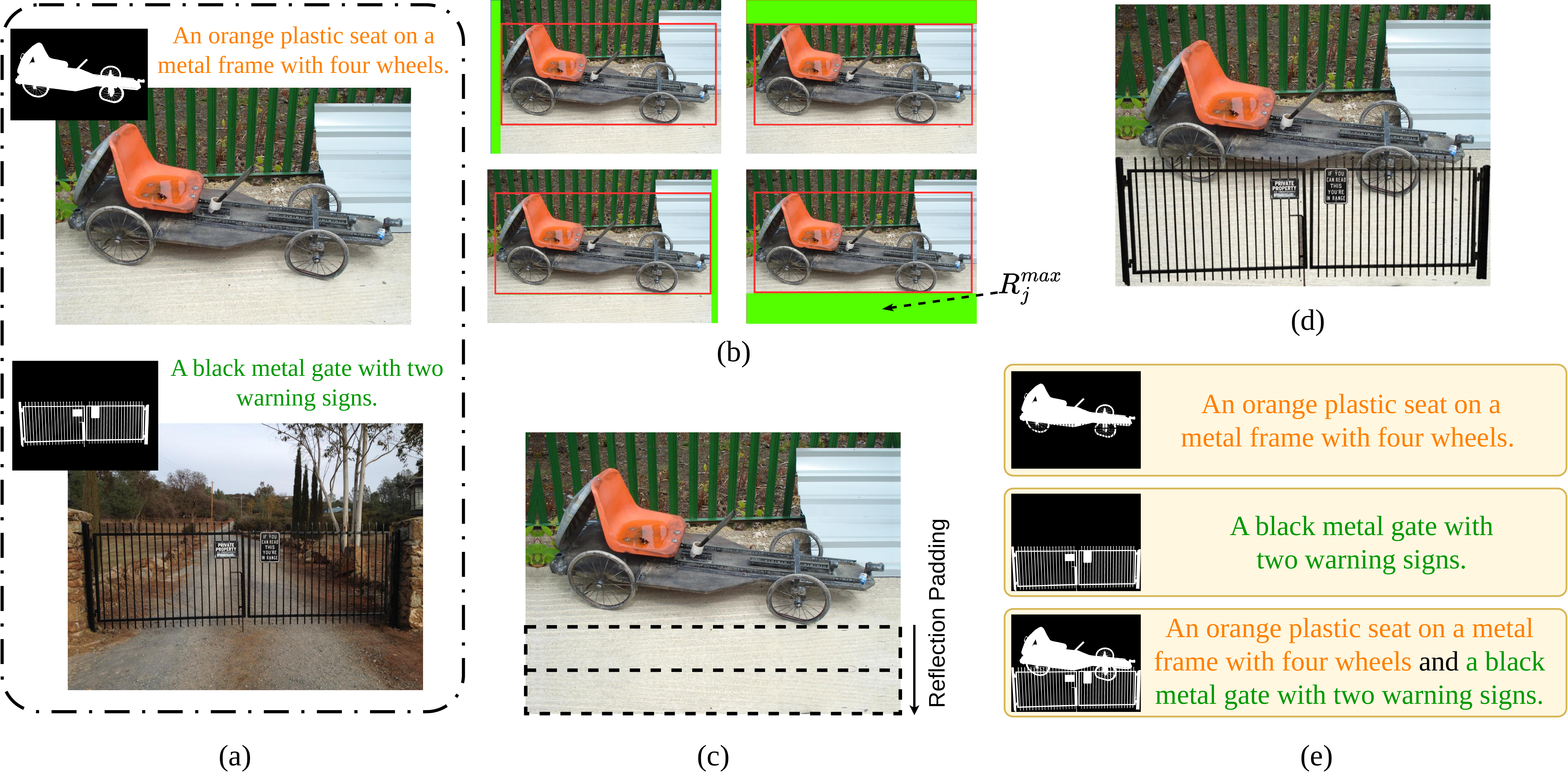}
    \caption{
    \textbf{Illustration of the Position-Aware Instance Pairing (PAIP) strategy.}
    \textbf{(a)} A reference sample (top triplet) is paired with another sample from the same batch (bottom triplet).
    \textbf{(b)} Candidate rectangular regions (in green) adjacent to the minimal bounding box (outlined in red) of the reference foreground are computed, and the one with the maximum area is selected as $R^{\text{max}}_j$.
    \textbf{(c)} The reference image is padded along the side adjacent to $R^{\text{max}}_j$ by an amount equal to the length of its opposite side.
    \textbf{(d)} The pairing foreground is cropped, resized, and placed within the designated placement area.
    \textbf{(e)} Mask and prompt options are then constructed, from which PAIP randomly selects one for training.
    }
    \label{fig:method:paip}
\end{figure*}

Although FlowDIS inherently supports text-conditioned segmentation, we introduce a Position-Aware Instance Pairing (PAIP) strategy to further enhance its language controllability by selectively pairing \textit{(image, mask, prompt)} triplets within each training mini-batch. This pairing encourages the model to learn from more diverse, multi-object scenes during training.  The overview of PAIP is illustrated in \cref{fig:method:paip}.

More precisely, given a batch of triplet instances $\{(I_j, M_j, \tau_j)\}_{j=1}^n$ of size $n$, for each triplet $(I_j, M_j, \tau_j)$ we randomly select another triplet $(I_k, M_k, \tau_k)$ with $k \neq j$. We refer to the former as the reference sample and the latter as the pairing sample.
We then combine the foreground of the pairing sample with the reference image $I_j$ to create a new image $I_{\text{mix}}$, which is then paired with a foreground mask $M_{\text{mix}}$ and prompt $\tau_{\text{mix}}$ to form the final training triplet $(I_{\text{mix}}, M_{\text{mix}}, \tau_{\text{mix}})$.

To construct the combined image $I_{mix}$, we first compute the minimal bounding box $B_j$ that spans the main object indicated by the reference mask $M_j$. 
We then find the largest rectangle in the image that is adjacent to $B_j$—i.e., it touches $B_j$ along one side but does not overlap (see \cref{fig:method:paip} (b))—and denote it as $R^{\text{max}}_j$. 
This region defines the space where we will place the pairing foreground object from $I_k$. 
In practice, $R^{\text{max}}_j$ is typically smaller than the bounding box $B_j$ of the reference foreground. 
To compensate for this, we enlarge the available background region by padding $I_j$ (and the corresponding mask $M_j$) along the side it shares with $R^{\text{max}}_j$.
The padding amount is set equal to the length of the opposite (non-shared) side of the rectangle $R^{\text{max}}_j$ (see \cref{fig:method:paip} (c)). 
Reflection padding is applied to preserve visual continuity and maintain a natural appearance. 
This operation effectively doubles the size of the initially selected placement region $R_j^{max}$ of $I_j$.
We denote the padded reference image as $I^{\text{pad}}_j$.

To combine the object of the pairing image $I_k$ with $I^{\text{pad}}_j$, we first crop its foreground region and resize it to fit within the placement area of the padded reference image $I^{\text{pad}}_j$, while preserving the original aspect ratio of the foreground.
The resized pairing foreground is then randomly placed within this area with minimal overlap with the foreground object of $I_j$. Alpha blending is used to ensure seamless integration with $I_j^{\text{pad}}$. As a result, we get the combined image $I_{\text{mix}}$ (see \cref{fig:method:paip} (d)).

In $I_{\text{mix}}$, two primary objects from the reference and pairing images are present, each associated with its corresponding binary mask.
Due to the padding, shifting, and resizing operations, these masks differ from the original masks $M_j$ and $M_k$, and we denote the resulting ones as $\hat{M}_j$ and $\hat{M}_k$.
Finally, the mask $M_{\text{mix}}$ is randomly selected from the following generated set of options: $\{ \hat{M}_j \; \text{AND} \; (\hat{M}_k)^c, \; \hat{M}_k, \; \hat{M}_j \; \text{OR} \; \hat{M}_k \}$, where $(\hat{M}_k)^c$ denotes the complement of $\hat{M}_k$, i.e. $(\hat{M}_k)^c = 1-\hat{M}_k$, AND denotes the pixel-wise multiplication, and OR denotes the pixel-wise maximum.
We then select the corresponding textual description $\tau_{\text{mix}}$ from $\{\tau_j, \tau_k, \text{``}\tau_j \text{ and } \tau_k\text{''}\}$ according to the chosen $M_{\mathrm{mix}}$ as shown in \cref{fig:method:paip} (e).

\section{Experiments}

\begin{table*}[ht]
\centering
\scriptsize
\renewcommand{\arraystretch}{1.2}
\renewcommand{\tabcolsep}{1.65mm}
\begin{tabular}{r|ccccccccccccccc}
& \multicolumn{5}{c|}{DIS-TE1 (500 images)} & \multicolumn{5}{c|}{DIS-TE2 (500 images)} & \multicolumn{5}{c}{DIS-TE3 (500 images)} \\ 
Methods &$F_\beta^\omega \uparrow$ &$F_\beta^{mx}\uparrow$ & $\mathcal{M} \downarrow$ & $\mathcal{S}_{\alpha} \uparrow$ & \multicolumn{1}{c|} {$E_\phi^{mn} \uparrow$}  & $F_\beta^\omega \uparrow$ &$F_\beta^{mx}\uparrow$  & $\mathcal{M} \downarrow$ & $\mathcal{S}_{\alpha} \uparrow$ & \multicolumn{1}{c|}{$E_\phi^{mn} \uparrow$}  & $F_\beta^\omega \uparrow$ &$F_\beta^{mx}\uparrow$  & $\mathcal{M} \downarrow$ & $\mathcal{S}_{\alpha} \uparrow$ & $E_\phi^{mn} \uparrow$  \\ \hline
IS-Net$_{22}$ \cite{qin2022highly} & 0.662 &0.740 & 0.074 & 0.787 & \multicolumn{1}{c|}{0.820}  & 0.728 &0.799 & 0.070 & 0.823 & \multicolumn{1}{c|}{0.858}  & 0.758 & 0.830 & 0.064 & 0.836 & 0.883  \\
InSPyReNet$_{22}$ \cite{kim2022revisiting} & 0.788 &0.845 & 0.043 & 0.873 & \multicolumn{1}{c|}{0.894}  & 0.846 &0.894 & 0.036 & 0.905 & \multicolumn{1}{c|}{0.928} &  0.871 &0.919  & 0.034 & 0.918 & 0.943  \\
FP-DIS$_{23}$ \cite{zhou2023dichotomous} & 0.713 &0.784  & 0.060 & 0.821 & \multicolumn{1}{c|}{0.860}  & 0.767 &0.827 & 0.059 & 0.845 & \multicolumn{1}{c|}{0.893}  & 0.811 &0.868  & 0.049 & 0.871 & 0.922  \\
UDUN$_{23}$ \cite{pei2023unite} & 0.720 &0.784 & 0.059 & 0.817 & \multicolumn{1}{c|}{0.864}  & 0.768 &0.829 & 0.058 & 0.843 & \multicolumn{1}{c|}{0.886} & 0.809 &0.865 & 0.050 & 0.865 & 0.917 \\
BiRefNet$_{24}$ \cite{zheng2024birefnet} & 0.820 &0.860  & 0.037 & 0.885 & \multicolumn{1}{c|}{0.912}  & 0.858 &0.893 & 0.036 & 0.900 & \multicolumn{1}{c|}{0.931}  & 0.894 &0.925 & 0.028 & 0.919 & 0.957  \\
GenPercept$_{24}$ \cite{xu2024diffusion} & 0.794 &0.844 & 0.038 & 0.871 & \multicolumn{1}{c|}{0.909} & 0.827 &0.875 & 0.040 & 0.887 & \multicolumn{1}{c|}{0.925} &  0.840 &0.890 & 0.039 & 0.893 & 0.939  \\
MVANet$_{24}$ \cite{yu2024multi} & 0.825 &0.873 & 0.037 & 0.887 & \multicolumn{1}{c|}{0.916}  & 0.879 &0.916 & 0.030 & 0.918 & \multicolumn{1}{c|}{0.943} & 0.891 &0.929 & 0.030 & 0.923 & 0.952  \\
DiffDIS$_{25}$ \cite{DiffDIS} & 0.820 & 0.895 & 0.035 & 0.900 & \multicolumn{1}{c|}{0.905} & 0.859 &0.923 & 0.032 & 0.922 & \multicolumn{1}{c|}{0.927} &  0.877 &0.940 & 0.032 & 0.929 & 0.936  \\
PDFNet$_{25}$ \cite{liu2025highprecisiondichotomousimagesegmentation} & 0.845 &0.888 & 0.031 & 0.887 & \multicolumn{1}{c|}{0.916}  & 0.884 &0.919 & 0.029 & 0.921 & \multicolumn{1}{c|}{0.946} & 0.888 &0.929 & 0.029 & 0.923 & 0.953  \\
LawDIS$_{25}$ \cite{yan2025lawdis} & 0.866 &0.899 & 0.029 & 0.906 & \multicolumn{1}{c|}{0.934} & 0.888 &0.921 & 0.030 & 0.920 & \multicolumn{1}{c|}{0.947} &  0.899 &0.929 & 0.028 & 0.924 & 0.955  \\
\mygray 
\textbf{Ours (1-step)} & \secondcolor{ 0.939} & \bestcolor{ 0.961} & \bestcolor{ 0.012} & \bestcolor{ \textbf{0.953}} & \multicolumn{1}{c|}{\secondcolor{ 0.975}}  & \secondcolor{ 0.944} & \bestcolor{ 0.965} & \secondcolor{ 0.013} & \bestcolor{ \textbf{0.958}} & \multicolumn{1}{c|}{\secondcolor{ 0.976}}  & \secondcolor{ 0.939} & \secondcolor{ 0.962} & \secondcolor{ 0.016} & \bestcolor{ 0.954} & \secondcolor{ 0.973}   \\
\mygray 
\textbf{Ours (2-step)} & \bestcolor{ \textbf{0.942}} & \bestcolor{ \textbf{0.961}} & \bestcolor{ \textbf{0.012}} & \bestcolor{ \textbf{0.953}} & \multicolumn{1}{c|}{\bestcolor{ \textbf{0.976}}}  & \bestcolor{ \textbf{0.947}} & \bestcolor{ \textbf{0.965}} & \bestcolor{ \textbf{0.012}} & \bestcolor{ \textbf{0.958}} & \multicolumn{1}{c|}{\bestcolor{ \textbf{0.978}}}  & \bestcolor{ \textbf{0.942}} & \bestcolor{ \textbf{0.963}} & \bestcolor{ \textbf{0.015}} & \bestcolor{ \textbf{0.954}} & \bestcolor{ \textbf{0.975}}  \\ \hline
& \multicolumn{5}{c|}{DIS-TE4 (500 images)} & \multicolumn{5}{c|}{DIS-TE (1-4) (2,000 images)} & \multicolumn{5}{c}{DIS-VD (470 images)} \\ 
Methods &$F_\beta^\omega \uparrow$ &$F_\beta^{mx}\uparrow$ & $\mathcal{M} \downarrow$ & $\mathcal{S}_{\alpha} \uparrow$ & \multicolumn{1}{c|}{$E_\phi^{mn} \uparrow$}  & $F_\beta^\omega \uparrow$ &$F_\beta^{mx}\uparrow$  & $\mathcal{M} \downarrow$ & $\mathcal{S}_{\alpha} \uparrow$ & \multicolumn{1}{c|}{$E_\phi^{mn} \uparrow$}  & $F_\beta^\omega \uparrow$ &$F_\beta^{mx}\uparrow$  & $\mathcal{M} \downarrow$ & $\mathcal{S}_{\alpha} \uparrow$ & $E_\phi^{mn} \uparrow$ \\ \hline
IS-Net$_{22}$ \cite{qin2022highly} & 0.753 & 0.827 & 0.072 & 0.830 & \multicolumn{1}{c|}{0.870}  & 0.726 & 0.799 & 0.070 & 0.819 & \multicolumn{1}{c|}{0.858}  & 0.717 & 0.791 & 0.074 & 0.813 & 0.856  \\
InSPyReNet$_{22}$ \cite{kim2022revisiting} & 0.848 & 0.905 & 0.042 & 0.905 & \multicolumn{1}{c|}{0.928}  & 0.838 & 0.891 & 0.039 & 0.900 & \multicolumn{1}{c|}{0.923}  & 0.834 & 0.889 & 0.042 & 0.900 & 0.922  \\
FP-DIS$_{23}$ \cite{zhou2023dichotomous} & 0.788 & 0.846 & 0.061 & 0.852 & \multicolumn{1}{c|}{0.906}  & 0.770 & 0.830 & 0.057 & 0.847 & \multicolumn{1}{c|}{0.895}  & 0.763 & 0.823 & 0.062 & 0.843 & 0.891  \\
UDUN$_{23}$ \cite{pei2023unite} & 0.792 & 0.846 & 0.059 & 0.849 & \multicolumn{1}{c|}{0.901}  & 0.772 & 0.831 & 0.057 & 0.844 & \multicolumn{1}{c|}{0.892}  & 0.763 & 0.823 & 0.059 & 0.838 & 0.892  \\
BiRefNet$_{24}$ \cite{zheng2024birefnet} & 0.865 & 0.904 & 0.039 & 0.900 & \multicolumn{1}{c|}{0.941}  & 0.858 & 0.896 & 0.035 & 0.901 & \multicolumn{1}{c|}{0.934}  & 0.855 & 0.891 & 0.038 & 0.898 & 0.932  \\
GenPercept$_{24}$ \cite{xu2024diffusion} & 0.801 & 0.861 & 0.055 & 0.869 & \multicolumn{1}{c|}{0.918}  & 0.816 & 0.868 & 0.043 & 0.880 & \multicolumn{1}{c|}{0.923}  & 0.815 & 0.865 & 0.043 & 0.881 & 0.922  \\
MVANet$_{24}$ \cite{yu2024multi} & 0.865 & 0.912 & 0.038 & 0.908 & \multicolumn{1}{c|}{0.939}  & 0.862 & 0.907 & 0.034 & 0.909 & \multicolumn{1}{c|}{0.938}  & 0.863 & 0.904 & 0.034 & 0.908 & 0.936  \\
DiffDIS$_{25}$ \cite{DiffDIS} & 0.835 &0.915 & 0.045 & 0.911 & \multicolumn{1}{c|}{0.910} & 0.848 &0.918 & 0.036 & 0.916 & \multicolumn{1}{c|}{0.919} &  0.844 &0.915 & 0.037 & 0.917 & 0.917  \\
PDFNet$_{25}$ \cite{liu2025highprecisiondichotomousimagesegmentation} & 0.867 &0.910 & 0.038 & 0.909 & \multicolumn{1}{c|}{0.941}  & 0.874 &0.913 & 0.031 & 0.913 & \multicolumn{1}{c|}{0.943} & 0.873 &0.913 & 0.030 & 0.915 & 0.944  \\
LawDIS$_{25}$ \cite{yan2025lawdis} & 0.884 &0.922 & 0.034 & 0.915 & \multicolumn{1}{c|}{0.952} & 0.884 &0.918 & 0.030 & 0.916 & \multicolumn{1}{c|}{0.947} &  0.884 &0.917 & 0.030 & 0.917 & 0.949  \\
\mygray 
\textbf{Ours (1-step)} & \secondcolor{ 0.912} & \secondcolor{ 0.945} & \secondcolor{ 0.026} & \secondcolor{ 0.938} & \multicolumn{1}{c|}{\secondcolor{ 0.961}}  & \secondcolor{ 0.933} & \secondcolor{ 0.958} & \secondcolor{ 0.017} & \bestcolor{ 0.951} & \multicolumn{1}{c|}{\secondcolor{ 0.971}}  & \secondcolor{ 0.933} & \secondcolor{ 0.957} & \secondcolor{ 0.015} & \secondcolor{ 0.952} & \secondcolor{0.972}  \\
\mygray 
\textbf{Ours (2-step)} & \bestcolor{ \textbf{0.919}} & \bestcolor{ \textbf{0.946}} & \bestcolor{ \textbf{0.024}} & \bestcolor{ \textbf{0.939}} & \multicolumn{1}{c|}{\bestcolor{ \textbf{0.964}}}  & \bestcolor{ \textbf{0.938}} & \bestcolor{ \textbf{0.959}} & \bestcolor{ \textbf{0.016}} & \bestcolor{ \textbf{0.951}} & \multicolumn{1}{c|}{\bestcolor{ \textbf{0.973}}}  & \bestcolor{ \textbf{0.938}} & \bestcolor{ \textbf{0.958}} & \bestcolor{ \textbf{0.014}} & \bestcolor{ \textbf{0.953}} & \bestcolor{ \textbf{0.974}}  \\  
\end{tabular}

\caption{Quantitative comparison with 10 representative methods on the DIS5K dataset. All results are evaluated at $1024 \times 1024$ px input resolution. $\downarrow$ indicates lower is better, while $\uparrow$ means higher is better. The best and second-best results are highlighted in \bestcolor{\textbf{bold}} and \secondcolor{underlined}, respectively.}
\label{tab:quantitative_sota}

\end{table*}

\begin{figure*}[h!]
    \centering
    \includegraphics[width=\linewidth]{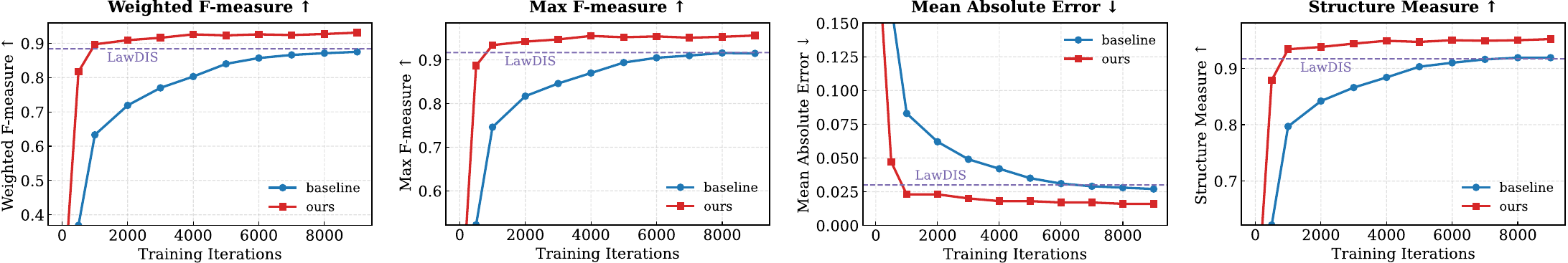}
    \caption{Metrics evaluated at different training iteration checkpoints for our method vs the baseline (image-conditioned mask generation from noise). Dashed lines show the metric levels of fully trained LawDIS \cite{yan2025lawdis}. Our FlowDIS converges faster than the baseline denoising approach and surpasses state-of-the-art LawDIS after only 1K training iterations, whereas LawDIS was trained for 36K iterations.}
    \label{fig:experiments:training_hist}
\end{figure*}

\subsection{Implementation Details}

\begin{algorithm}[t]
\caption{FlowDIS Inference}
\label{alg:flowdis-euler-beta}
\begin{algorithmic}[1]
% Redefine \COMMENT to be gray
\renewcommand{\COMMENT}[1]{\hfill\textcolor{gray}{\# #1}}
\REQUIRE v-pred. model $v_\theta$, VAE $(\mathcal{E}, \mathcal{D})$, image $I$, text condition $c_{\tau}$, steps $N \in \mathbb{N}$, $\alpha,\beta \in \mathbb{R^+}$
\STATE $q \gets \text{linspace}(0,1,N+1)$
\STATE $t_i \gets F_{\text{Beta}}^{-1}(q_i;\alpha,\beta), \quad i=0,\dots,N$ \COMMENT{Beta schedule}
\STATE $\hat{z}_{t_N} = z^I \gets \mathcal{E}(I)$ \COMMENT{Encode image into latent space}
\FOR{$n = N-1, \dots, 0$}
    \STATE $\hat{z}_{t_n} \gets \hat{z}_{t_{n+1}} + v_\theta(\hat{z}_{t_{n+1}}, z^I, t_{n+1}, c_{\tau}) (t_n - t_{n+1})$
\ENDFOR
\STATE $\hat{M}_{\text{rgb}} \gets \mathcal{D}(\hat{z}_0)$ \COMMENT{Decode latent to RGB mask}
\STATE $\hat{M}_{i,j} \gets \frac{1}{3} \sum_{k=1}^{3} (\hat{M}_{\text{rgb}})_{i,j,k}$ \COMMENT{Convert to grayscale}
\STATE $\hat{M} \leftarrow \mathrm{clip}(\hat{M}, 0, 1)$
\RETURN $\hat{M}$ \COMMENT{Final predicted mask}
\end{algorithmic}
\end{algorithm}

As the base flow matching MMDiT model, we adopt FLUX.1-Schnell \cite{flux2024}, initialized with its pre-trained weights.
To incorporate an additional image latent condition, we extend the input channels of the first linear layer in the transformer, initializing the new weights to zeros. For text guidance, CLIP \cite{radford2021learning} and T5 \cite{raffel2020exploring} are used as text encoders to obtain the token sequence $c_{\tau}$. 
During training, the timestep distribution $p(t)$ is a $\mathrm{Beta}(2.5, 1)$ distribution, which biases the training toward larger $t$ values, where prediction is more challenging. 
The models are trained with a batch size of 32 for 10,000 iterations ($\approx1.8$ days) on 8 $\times$ NVIDIA A100 GPUs. For optimization, we use the AdamW optimizer with an initial learning rate of $5\times 10^{-5}$, which is halved at steps 512, 2048, 4096, and 8192.

For inference, we adopt a non-uniform schedule derived from the Beta cumulative distribution function (CDF).
% \SN{we should cite the reason of the RESOLUTION-DEPENDENT SHIFT (just citing the paper is enough)}
Given $N$ total timesteps, we first define an equidistant grid $q \in [0, 1]$ of length $N+1$, which is then mapped using the inverse Beta CDF:
\begin{equation}
t_i = F^{-1}_{\text{Beta}}(q_i; \alpha, \beta), \quad i = 0, \dots, N,
\end{equation}
where $F^{-1}_{\text{Beta}}(\cdot; \alpha, \beta)$ denotes the inverse CDF with shape parameters $\alpha, \beta \in \mathbb{R}^{+}$. This schedule enables denser sampling near the start or end of the flow trajectory depending on $(\alpha, \beta)$, providing finer control over integration dynamics. We use $\alpha = 2.5$ and $\beta = 1.0$, consistent with training. Our full inference pipeline is summarized in Algorithm~\ref{alg:flowdis-euler-beta}.

\begin{figure*}[h!]
    \centering
    \includegraphics[width=\linewidth]{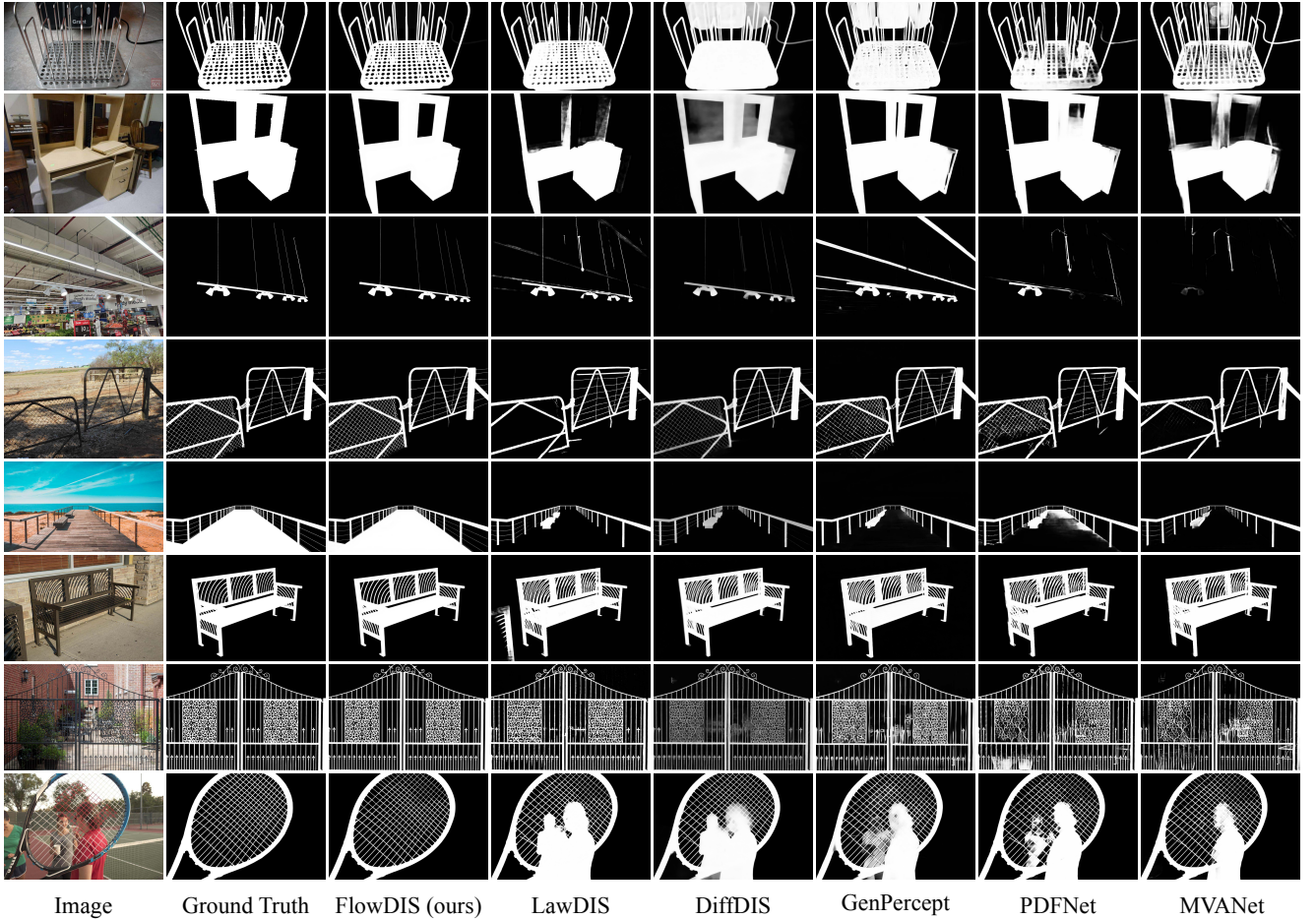}
    \caption{Qualitative comparison with state-of-the-art DIS methods. Our approach generates more detailed and more semantically accurate masks.}
    \label{fig:experiments:qualitative}
\end{figure*}

\subsection{Experimental Setup}
\label{sec:experiments:setup}
\textbf{Dataset:} We conduct our experiments on the DIS5K \cite{qin2022highly} dataset, which consists of 5,470 high-resolution image-mask pairs across 225 categories.
It is further divided into DIS-TR (3,000 images), DIS-VD (470 images), and DIS-TE (2,000 images). 
All training experiments are conducted on DIS-TR, while DIS-VD and DIS-TE are exclusively used for testing. 
For language guidance, we generate captions $\tau$ using a vision-language model. We discuss this process in detail in the appendix.

DIS-TE is further divided into DIS-TE1, DIS-TE2, DIS-TE3, and DIS-TE4 subsets, each containing 500 images, where the numbers 1–4 denote increasing levels of foreground complexity. 
We test our approach on all subsets separately, as well as on the combined test set DIS-TE (1-4), to compare FlowDIS across all complexity levels. 

\noindent \textbf{Evaluation metrics:} We use widely adopted metrics: the weighted F-measure ($F_\beta^\omega \uparrow$) \cite{6909433}, max F-measure ($F_\beta^{mx} \uparrow$) \cite{perazzi2012saliency}, mean absolute error ($\mathcal{M} \downarrow$) \cite{perazzi2012saliency}, Structure-measure ($S_{\alpha} \uparrow$) \cite{cheng2021structure} and E-measure ($E_\phi^{mn} \uparrow$) \cite{Fan2018Enhanced}.

\subsection{Quantitative and Qualitative Analysis}

\begin{figure*}
    \centering
    \includegraphics[width=\linewidth]{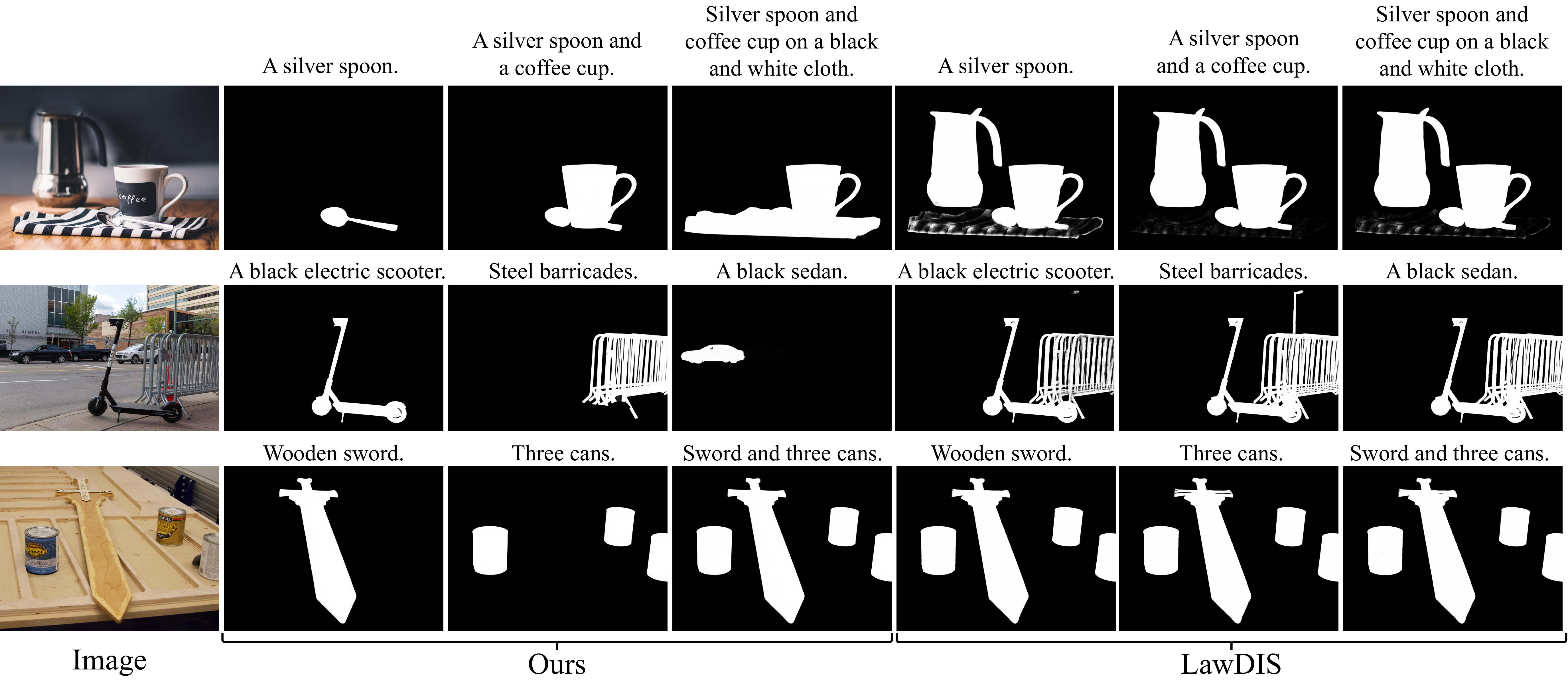}
    \caption{Comparison of language controllability between our method and LawDIS \cite{yan2025lawdis}. Each output is generated using the corresponding text prompt shown above.}
    \label{fig:experiments:lang_guidance}
\end{figure*}

\label{sec:experiments:quantitative}

\textbf{Quantitative comparison.} We compare our FlowDIS with the following state-of-the-art DIS methods: IS-Net \cite{qin2022highly}, InSPyReNet \cite{kim2022revisiting}, FP-DIS \cite{zhou2023dichotomous}, UDUN \cite{pei2023unite}, BiRefNet \cite{zheng2024birefnet}, GenPercept \cite{xu2024diffusion}, MVANet \cite{yu2024multi}, DiffDIS \cite{DiffDIS}, PDFNet \cite{liu2025highprecisiondichotomousimagesegmentation} and LawDIS \cite{yan2025lawdis}. For a fair comparison, all methods are evaluated at an input resolution of $1024 \times 1024$ px. As shown in \cref{tab:quantitative_sota}, our method already achieves state-of-the-art results across all test sets with 1-step inference, while using 2-step inference yields even better performance. For example, on the full DIS test set, 1-step FlowDIS yields an $\approx5.5$\% improvement in $F^{\omega}_{\beta}$ and an $\approx43$\% reduction in $\mathcal{M}$ over the runner-up language-guided model LawDIS \cite{yan2025lawdis}.

\noindent \textbf{Qualitative comparison.} \cref{fig:experiments:qualitative} presents a qualitative comparison between FlowDIS and other state-of-the-art methods. As shown in rows 1, 4, and 7, our method captures fine-grained visual details more effectively, while in rows 2, 3, and 5, it demonstrates superior semantic understanding of the scene.

In \cref{fig:experiments:lang_guidance}, we compare the language controllability of our method with the state-of-the-art language-guided approach, LawDIS \cite{yan2025lawdis}. As shown, our method exhibits stronger controllability while simultaneously producing more accurate results. For instance, in the second row, our method correctly isolates the black sedan from a background containing two other cars, whereas LawDIS produces nearly identical outputs regardless of the input prompt. 
Notably, as can be seen from our qualitative results, while the introduced PAIP strategy constructs samples with a pair of foregrounds, FlowDIS generalizes to more complex scenes by successfully selecting the requested object.
More qualitative results can be found in the appendix.

\subsection{Ablation Studies}

We conduct ablation studies on the DIS-VD subset using fixed 2-step inference unless specified otherwise.

\noindent
\textbf{Effectiveness of our deterministic flow matching formulation:} We evaluate the effectiveness of the flow matching setup described in \cref{sec:method:flow_dis}. Specifically, we compare our approach (deterministic FM), which uses the image latent as $z_1$, against a variant where $z_1$ is sampled from standard Gaussian noise\footnote{We still provide $z^I$ to the MMDiT $v_\theta$ via concatenation to the input.} during training and inference (denoising FM). 
As can be seen from \cref{tab:ablation_fm}, our deterministic formulation significantly outperforms the denoising-based variant.

\begin{table}[H]
\centering
\vspace{-3pt}
\scriptsize
\renewcommand{\arraystretch}{1.2} % vertical spacing
\begin{tabular}{c|llllll}
Ablation settings &$F_\beta^\omega \uparrow$ &$F_\beta^{mx}\uparrow$ & $\mathcal{M} \downarrow$ & $\mathcal{S}_{\alpha} \uparrow$ &  $E_\phi^{mn} \uparrow$ \\ \hline
denoising FM  & 0.883  & 0.916 & 0.025 & 0.920 & 0.957 \\
deterministic FM &  \textbf{0.938}  & \textbf{0.958}  & \textbf{0.014} & \textbf{0.953} & \textbf{0.974} \\
\end{tabular}
\caption{Ablation study on flow matching (FM) settings.}
\vspace{-6pt}
\label{tab:ablation_fm}
\end{table}

We also compare the training convergence of the baseline denoising-based flow matching approach with our FlowDIS formulation by evaluating the models at different training iterations and plotting their performance. As shown in \cref{fig:experiments:training_hist}, our formulation converges significantly faster than the baseline. Moreover, our method requires only 1K training iterations to surpass the second-best model, LawDIS~\cite{yan2025lawdis}, while LawDIS was trained for 36K iterations.

\noindent
\textbf{Effectiveness of language guidance:} We train and evaluate FlowDIS with and without language prompts. As shown in \cref{tab:ablation_prompts}, incorporating language guidance significantly improves performance by providing semantic cues that help the model resolve ambiguous inputs.

\begin{table}[H]
\centering
\vspace{-3pt}
\scriptsize
\renewcommand{\arraystretch}{1.2} % vertical spacing
\begin{tabular}{c|llllll}
Ablation settings &$F_\beta^\omega \uparrow$ &$F_\beta^{mx}\uparrow$ & $\mathcal{M} \downarrow$ & $\mathcal{S}_{\alpha} \uparrow$ &  $E_\phi^{mn} \uparrow$ \\ \hline
\textbf{w/o} language guidance & 0.901 & 0.926  & 0.027 & 0.929 & 0.951 \\
\textbf{w/} language guidance & \textbf{0.937} & \textbf{0.956}  & \textbf{0.015}  & \textbf{0.952} & \textbf{0.975} \\
\end{tabular}
\caption{Ablation study on language guidance.}
\vspace{-6pt}
\label{tab:ablation_prompts}

\end{table}

\noindent
\textbf{Effectiveness of PAIP:} 
Since PAIP is intended to improve language controllability and the DIS-VD validation set contains mainly single objects, we construct a new benchmark by applying PAIP to the DIS-VD subset.
The resulting test set, DIS-VD-Complex, maintains the same number of samples as DIS-VD but introduces more complex visual compositions, which can be challenging for models with weaker language-following capabilities. 
As shown in \cref{tab:ablation_paip}, PAIP significantly improves the results on the complex-scene test set DIS-VD-Complex, while preserving performance on the simple-scene set DIS-VD.

\begin{table}[H]
\centering
\scriptsize
\setlength{\tabcolsep}{4pt} % horizontal spacing
\renewcommand{\arraystretch}{1.2} % vertical spacing
\begin{tabular}{c|lll|lll}
 & \multicolumn{3}{c|}{DIS-VD-Complex} & \multicolumn{3}{c}{DIS-VD} \\ 
\hline
Ablation settings & $F_\beta^{mx}\!\uparrow$ & $\mathcal{M}\!\downarrow$ & $\mathcal{S}_{\alpha}\!\uparrow$ & $F_\beta^{mx}\!\uparrow$ & $\mathcal{M}\!\downarrow$ & $\mathcal{S}_{\alpha}\!\uparrow$ \\ 
\hline
FlowDIS \textbf{w/o} PAIP & 0.783 & 0.063 & 0.831 & 0.956 & 0.015 & 0.952 \\
FlowDIS \textbf{w/} PAIP  & \textbf{0.960} & \textbf{0.014} & \textbf{0.955} & \textbf{0.958} & \textbf{0.014} & \textbf{0.953} \\
\end{tabular}
\caption{Ablation study on PAIP.}
\vspace{-6pt}
\label{tab:ablation_paip}
\end{table}

Additional quantitative and qualitative ablation results on PAIP are provided in the appendix.
\section{Conclusion}

We presented a novel, language-guided dichotomous image segmentation approach, \textbf{\textit{FlowDIS}}, which reformulates the segmentation task within the flow matching framework. Our formulation connects the predictive image segmentation task with generative modeling, while preserving the deterministic nature of segmentation. To further improve language controllability, we introduced a \textbf{\textit{Position-Aware Instance Pairing (PAIP)}} strategy, which constructs pairwise foreground compositions within each training batch while selecting the guidance prompt and the target mask. Extensive quantitative and qualitative experiments show that our approach surpasses existing state-of-the-art methods.

{
    \small
    \bibliographystyle{ieeenat_fullname}
    \bibliography{main}
}

% Appendix - comment before main paper submission %
\clearpage
\appendix
% \clearpage
% \maketitlesupplementary
\section*{\Large Appendix}
\section{Language-Prompt Generation Details}
\label{sec:supp:lang_prompt_details}

\begin{figure*}[ht!]
    \centering
    \includegraphics[width=\linewidth]{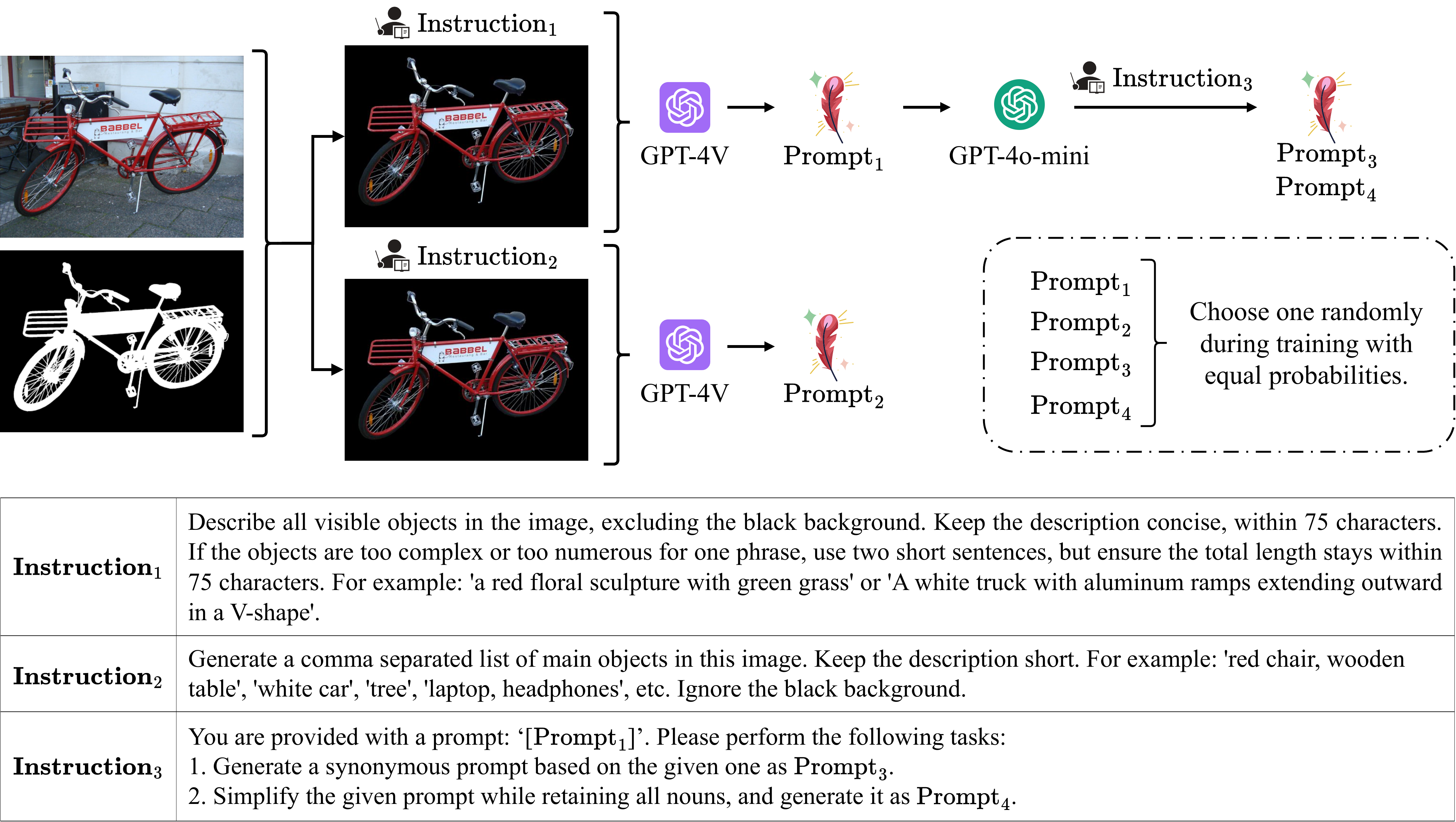}
    \caption{\textbf{Illustration of our language-prompt generation pipeline.} GPT-4V generates two prompts from the blacked-out background: a detailed prompt ($\text{Prompt}_1$) and a shorter prompt ($\text{Prompt}_2$). GPT-4o-mini then produces two paraphrased variants of the detailed prompt ($\text{Prompt}_3$ and $\text{Prompt}_4$). During training, one of the four prompts is uniformly sampled; at test time, $\text{Prompt}_1$ is used for determinism.}
    \vspace{0.3cm}
    \label{fig:supp:lang_prompt_pipe}
\end{figure*}

For language-prompt generation, we follow LawDIS \cite{yan2025lawdis} while further simplifying and enhancing their pipeline. \cref{fig:supp:lang_prompt_pipe} illustrates the overall process. Specifically, we use the multi-modal language model GPT-4V \cite{achiam2023gpt} to generate two prompt types from the blacked-out background: a relatively detailed prompt ($\text{Prompt}_1$) and a shorter prompt ($\text{Prompt}_2$). We then employ GPT-4o-mini \cite{hurst2024gpt} to produce two additional paraphrased variants of the detailed prompt, denoted as $\text{Prompt}_3$ and $\text{Prompt}_4$. During training, we uniformly sample one of these four prompts to increase linguistic diversity. For quantitative comparisons at test time, we always use $\text{Prompt}_1$ to ensure determinism. Through manual inspection, we found that our language-prompt generation method achieves better alignment between the foreground and the corresponding language description. Therefore, in all quantitative comparisons with LawDIS, we used our prompts to ensure a fair evaluation.

\section{Ablation for $z^I$ Conditioning}
\label{sec:supp:zI_conditioning}

We condition the velocity prediction model by concatenating $z^I$ to its input, enabling access to the input image at intermediate denoising steps. This design improves fine segmentation details during multi-step inference. To validate its effect, we perform an ablation study on DIS-TE (1--4) with 2-step inference. As shown in \cref{tab:supp:abl_z_I_concat}, this conditioning consistently improves all evaluation metrics.

\begin{table}[H]
\centering
\resizebox{\columnwidth}{!}{%
\scriptsize
\renewcommand{\arraystretch}{1.2} % vertical spacing
\begin{tabular}{c|lllll}
Method & $F_\beta^\omega \uparrow$ & $F_\beta^{mx}\uparrow$ & $\mathcal{M} \downarrow$ & $\mathcal{S}_{\alpha} \uparrow$ & $E_\phi^{mn} \uparrow$ \\ \hline
Ours \textbf{w/o} additional $z^I$ condition & 0.933 & 0.954 & 0.017 & 0.948 & 0.972 \\
 Ours (i.e. \textbf{w/} $z^I$ channel-wise concat) & \textbf{0.938} & \textbf{0.959} & \textbf{0.016} & \textbf{0.951} & \textbf{0.973}
 \\
\end{tabular}
}

\caption{Ablation for $z^I$ conditioning on DIS-TE (1-4).}
\label{tab:supp:abl_z_I_concat}
\end{table}

\section{Further Ablation on PAIP}
\label{sec:supp:further_paip_ablation}
To further evaluate the effect of PAIP, we build a test set from the COCO \cite{caesar2018coco} validation set, excluding stuff categories. We convert the instance annotations into semantic masks, resulting in 4,952 images with 14,246 binary masks, each corresponding to a semantic class. For each mask, we generate a text prompt using the pipeline described in \cref{sec:supp:lang_prompt_details}. The ablation results on this dataset are shown in \cref{tab:supp:paip_ablation_on_coco}. As shown in the table, PAIP consistently improves all metrics, indicating better language controllability.

\begin{table}[H]
\centering
\resizebox{\columnwidth}{!}{%
\scriptsize
\renewcommand{\arraystretch}{1.2} % vertical spacing
\begin{tabular}{c|lllll}
Method & $F_\beta^\omega \uparrow$ & $F_\beta^{mx}\uparrow$ & $\mathcal{M} \downarrow$ & $\mathcal{S}_{\alpha} \uparrow$ & $E_\phi^{mn} \uparrow$ \\ \hline
FlowDIS \textbf{w/o} PAIP & 0.327 & 0.351 & 0.191 & 0.561 & 0.542 \\
FlowDIS \textbf{w/} PAIP & \textbf{0.511} & \textbf{0.545} & \textbf{0.075} & \textbf{0.700} & \textbf{0.719} \\
\end{tabular}
}
\caption{Zero-shot ablation of PAIP on COCO-Object.}
\label{tab:supp:paip_ablation_on_coco}
\end{table}

\begin{figure*}[h!]
    \centering
    \includegraphics[width=\linewidth]{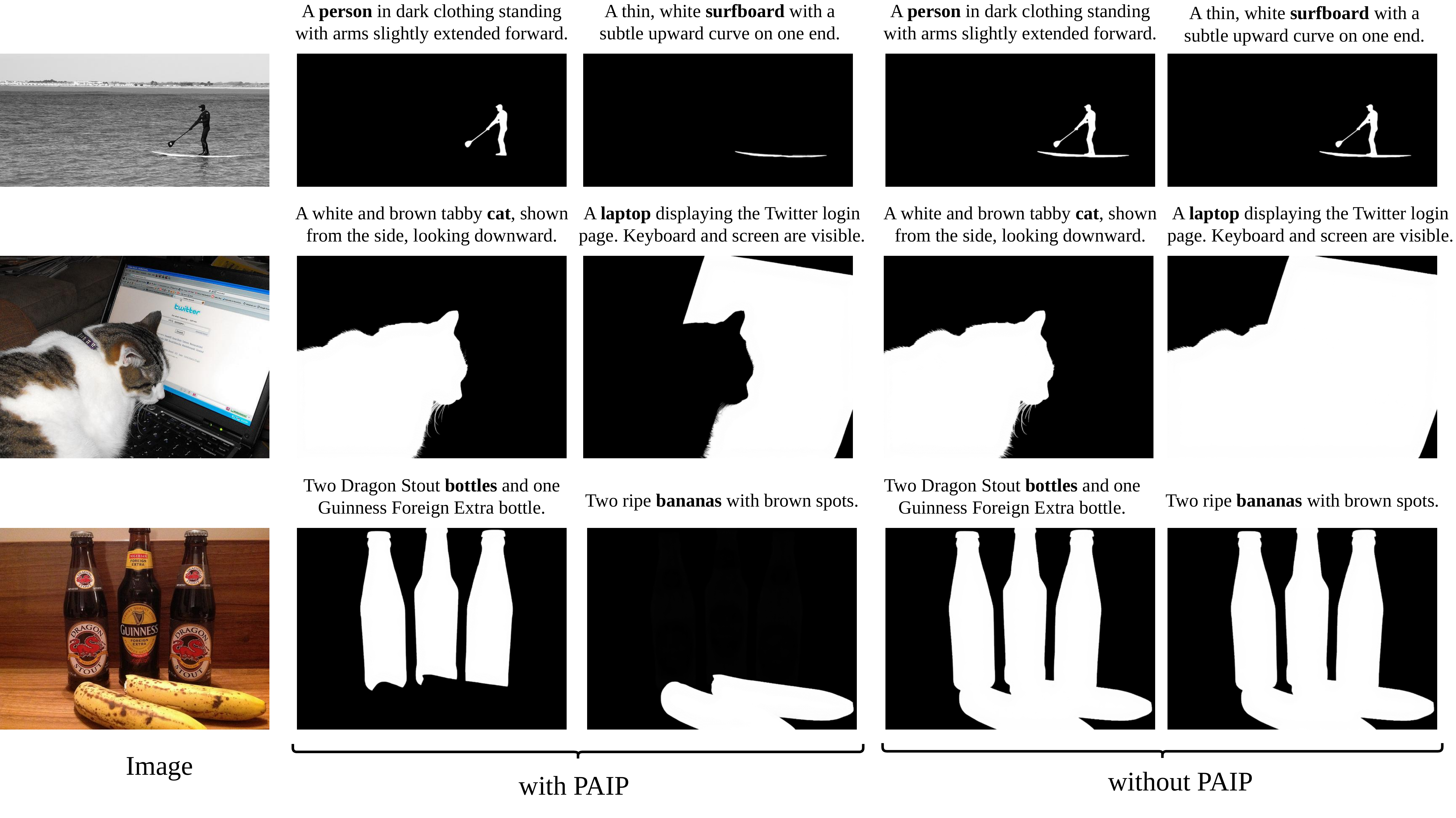}
    \caption{Qualitative ablation study of PAIP on samples from the COCO dataset.}
    \label{fig:supp:paip_qualitative}
\end{figure*}

These quantitative improvements are also reflected in the qualitative examples shown in \cref{fig:supp:paip_qualitative}. While FlowDIS without PAIP can struggle to follow the text prompt, the version with PAIP produces masks that better match the text description.

\section{Comparison with Open-Vocabulary Semantic Segmentation Methods}

Using the test set constructed in \cref{sec:supp:further_paip_ablation}, we compare FlowDIS with several state-of-the-art open-vocabulary semantic segmentation methods. In the zero-shot setting, FlowDIS achieves the best performance among the compared methods (see \cref{tab:supp:open_vocab_comparison}).

\begin{table}[H]
\centering
\resizebox{\columnwidth}{!}{%
\scriptsize
\renewcommand{\arraystretch}{1.2} % vertical spacing
\begin{tabular}{c|clllll}
Method & FlowDIS (Ours) & RF-CLIP \cite{li2025target111} & SCLIP \cite{wang2024sclip111} & FreeCP \cite{chen2025training111}  \\ \hline
mIoU $\uparrow$  & \textbf{47.7\%} & 31.8\% & 28.3\% & 21.6\%
\end{tabular}
}
\caption{Comparison with open-vocabulary semantic segmentation methods.}
\label{tab:supp:open_vocab_comparison}
\end{table}

\section{More Qualitative Comparisons}
\label{sec:more_qual_results}

\begin{figure*}[h!]
    \centering
    \includegraphics[width=\linewidth]{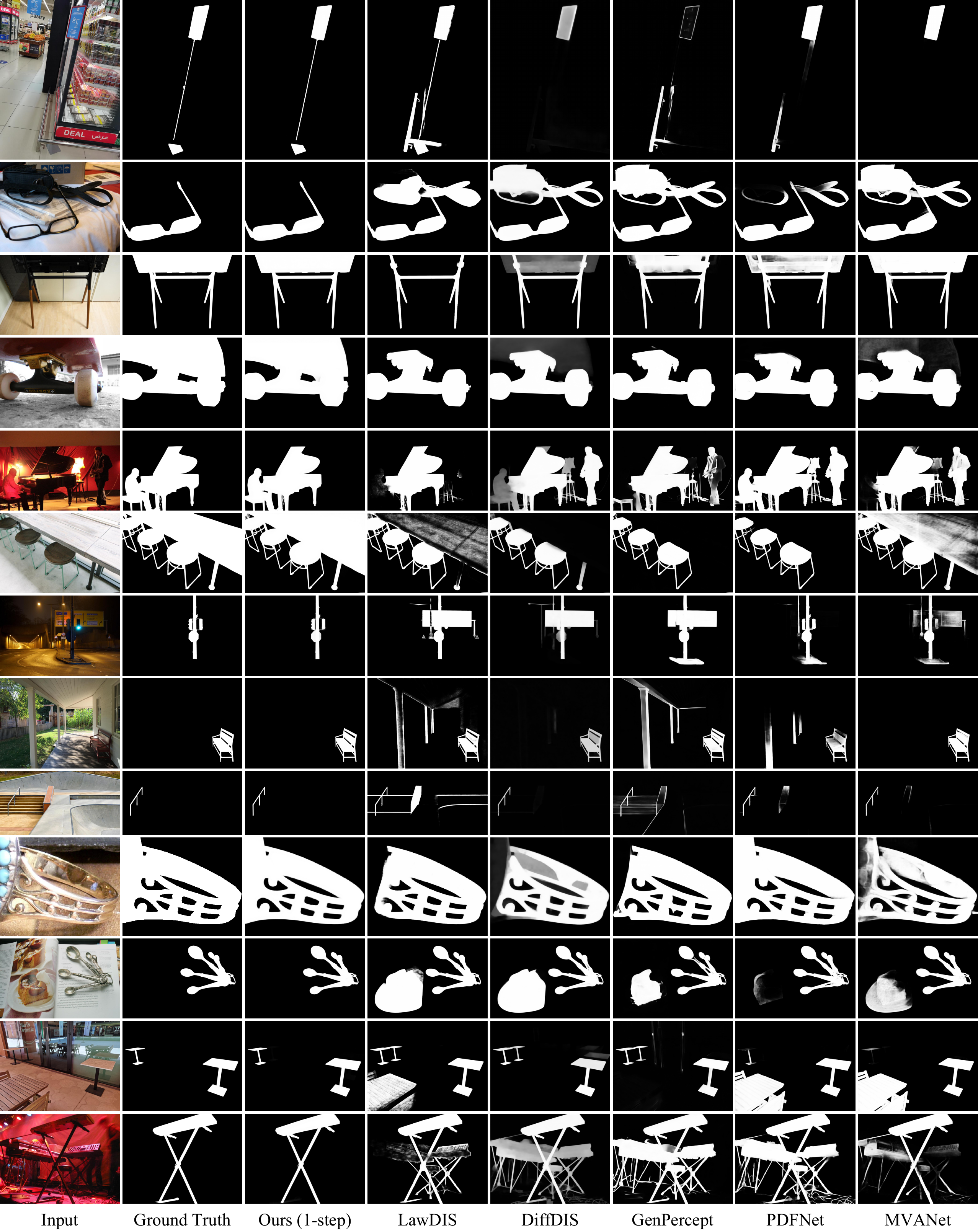}
    \caption{Qualitative comparison with state-of-the-art DIS methods on DIS-TE1. Please zoom in to compare finer details.}
    \label{fig:supp:qual_dis_te1}
\end{figure*}

\begin{figure*}[h!]
    \centering
    \includegraphics[width=\linewidth]{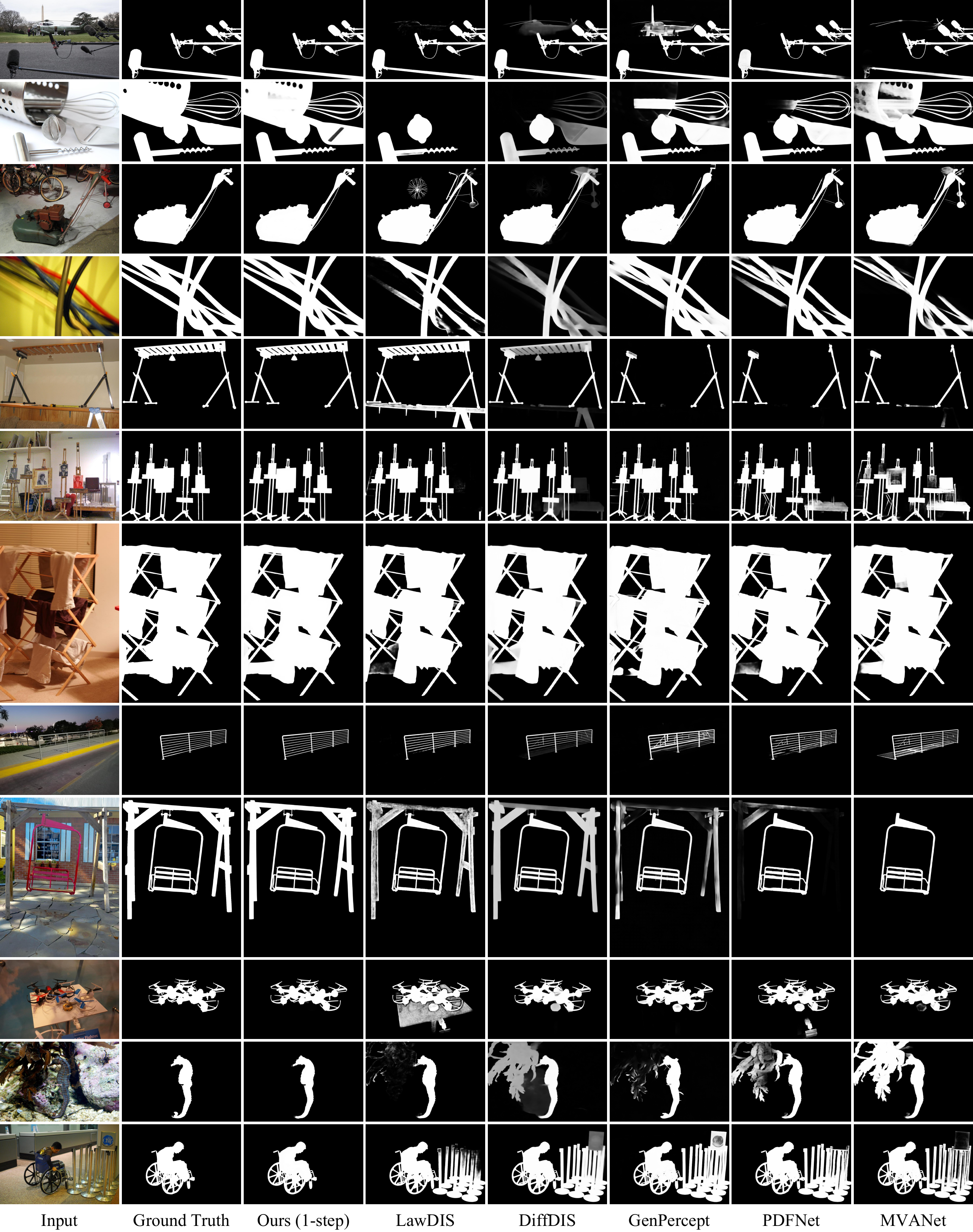}
    \caption{Qualitative comparison with state-of-the-art DIS methods on DIS-TE2. Please zoom in to compare finer details.}
    \label{fig:supp:qual_dis_te2}
\end{figure*}

\begin{figure*}[h!]
    \centering
    \includegraphics[width=\linewidth]{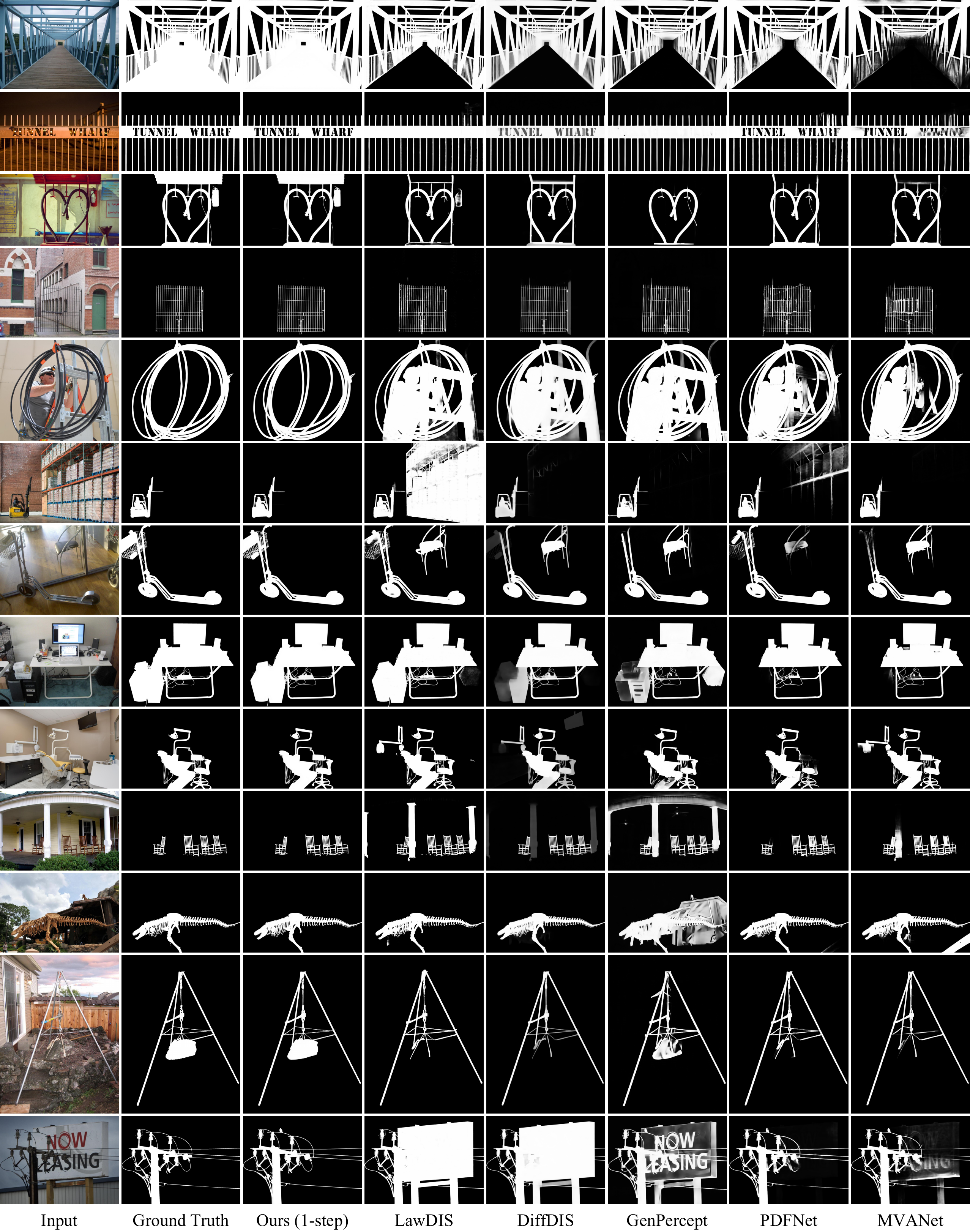}
    \caption{Qualitative comparison with state-of-the-art DIS methods on DIS-TE3. Please zoom in to compare finer details.}
    \label{fig:supp:qual_dis_te3}
\end{figure*}

\begin{figure*}[h!]
    \centering
    \includegraphics[width=\linewidth]{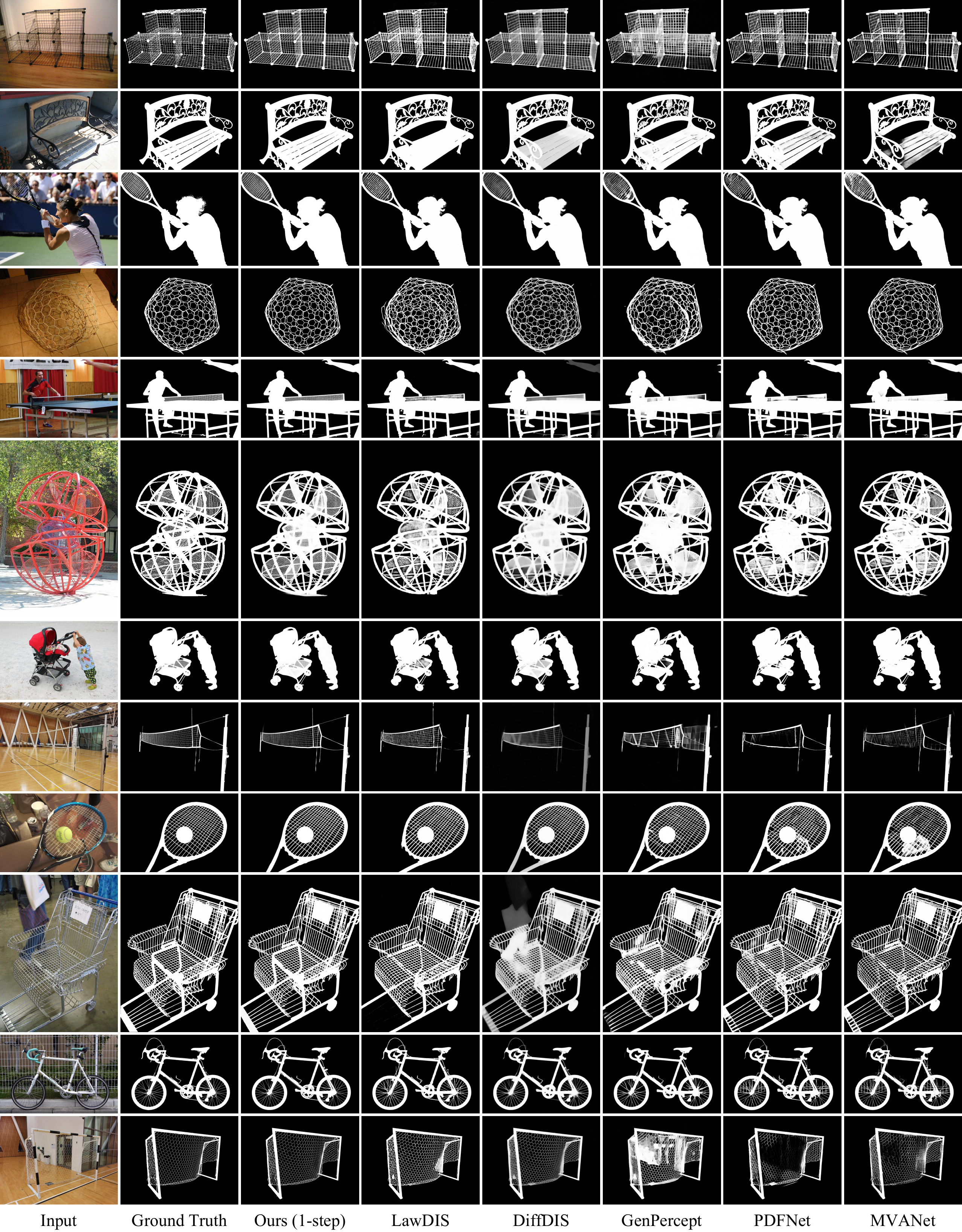}
    \caption{Qualitative comparison with state-of-the-art DIS methods on DIS-TE4. Please zoom in to compare finer details.}
    \label{fig:supp:qual_dis_te4}
\end{figure*}

\begin{figure*}[h!]
    \centering
    \includegraphics[width=\linewidth]{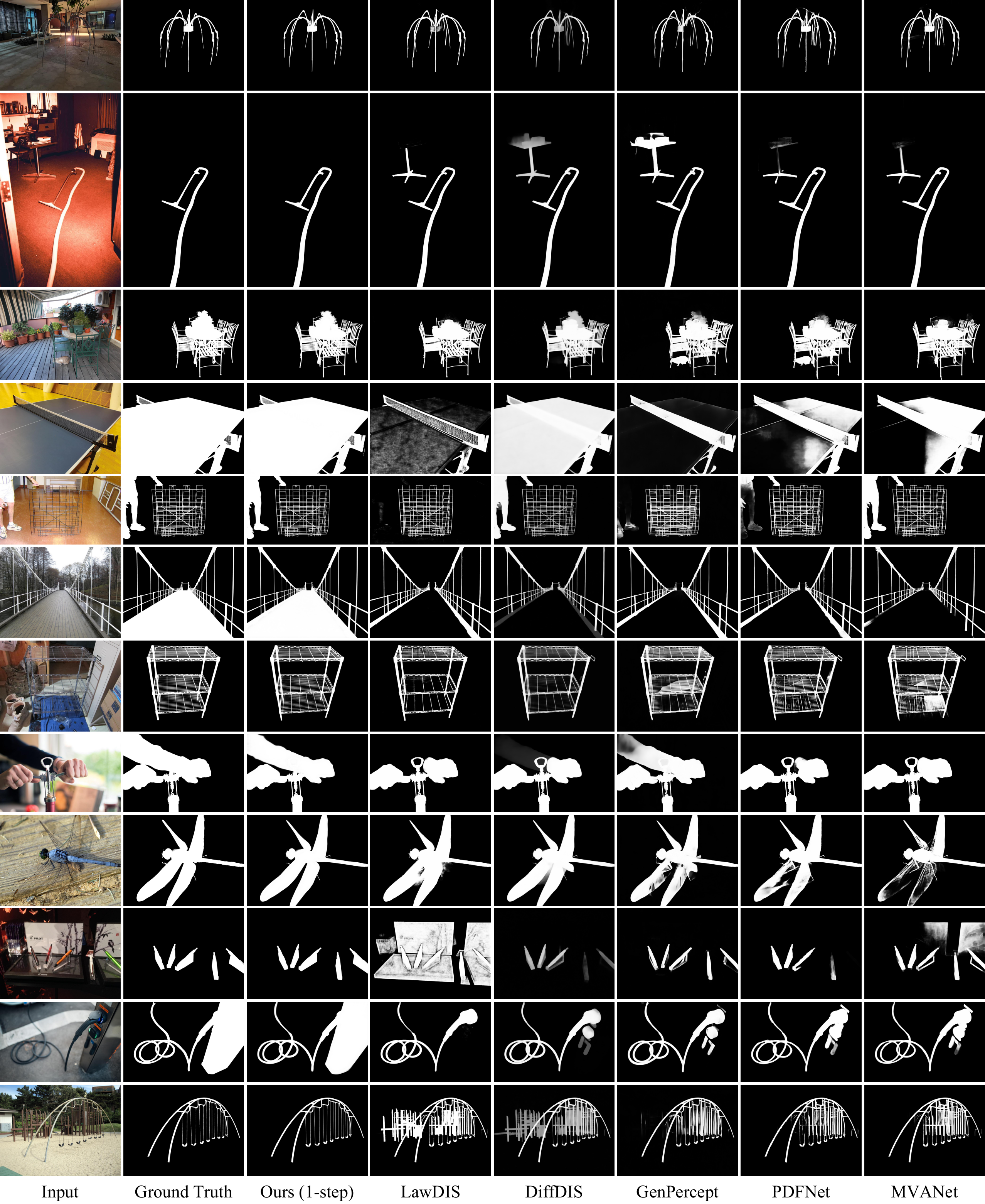}
    \caption{Qualitative comparison with state-of-the-art DIS methods on DIS-VD. Please zoom in to compare finer details.}
    \label{fig:supp:qual_dis_vd}
\end{figure*}

For a more comprehensive qualitative comparison, we compare our single-step results with other state-of-the-art methods on additional samples from the DIS5K test sets: DIS-TE1 (see \cref{fig:supp:qual_dis_te1}), DIS-TE2 (see \cref{fig:supp:qual_dis_te2}), DIS-TE3 (see \cref{fig:supp:qual_dis_te3}), DIS-TE4 (see \cref{fig:supp:qual_dis_te4}), and DIS-VD (see \cref{fig:supp:qual_dis_vd}). All results are generated with 1-step inference at $1024 \times 1024$ px resolution for a fair comparison.

\begin{figure*}[h!]
    \centering
    \includegraphics[width=\linewidth]{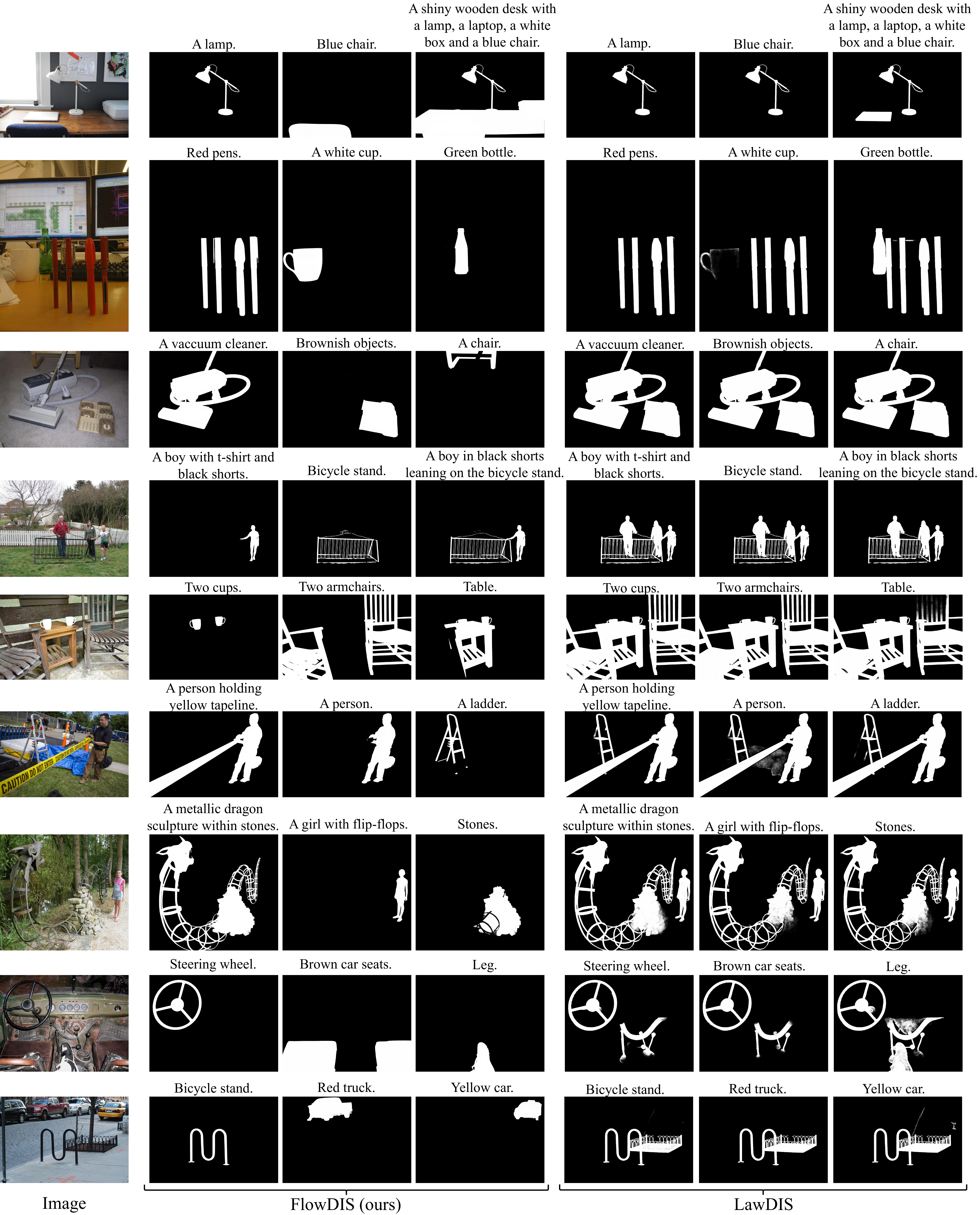}
    \caption{Comparison of language controllability between our FlowDIS and LawDIS \cite{yan2025lawdis}. Each output is generated using the corresponding text prompt shown above.}
    \label{fig:supp:qual_lang_guidance}
\end{figure*}

\cref{fig:supp:qual_lang_guidance} shows additional samples demonstrating the language controllability of FlowDIS compared with the state-of-the-art method LawDIS \cite{yan2025lawdis}.

\section{Resolution Scaling}
\label{sec:resolution_scaling}

\begin{table}[H]
\centering
\resizebox{\columnwidth}{!}{%
\scriptsize
\renewcommand{\arraystretch}{1.2} % vertical spacing
\begin{tabular}{c|lllll}
Inference res. & $F_\beta^\omega \uparrow$ & $F_\beta^{mx}\uparrow$ & $\mathcal{M} \downarrow$ & $\mathcal{S}_{\alpha} \uparrow$ & $E_\phi^{mn} \uparrow$ \\ \hline
1024 $\times$ 1024 px & 0.919 & 0.946 & 0.024 & 0.939 & 0.964 \\
1280 $\times$ 1280 px & 0.925 & 0.952 & 0.023 & 0.944 & 0.965 \\
1536 $\times$ 1536 px & 0.928 & 0.953  & 0.022  & 0.945 & 0.966 \\
1792 $\times$ 1792 px & 0.931 & 0.955  & 0.022  & 0.946 & 0.966 \\
2048 $\times$ 2048 px & \textbf{0.932} & \textbf{0.956}  & \textbf{0.021}  & \textbf{0.947} & \textbf{0.967} \\
\end{tabular}
}
\caption{Performance metrics of FlowDIS at different input resolutions on DIS-TE4.}
\label{tab:supp:resolution}
\end{table}

\begin{figure*}[h!]
    \centering
    \includegraphics[width=\linewidth]{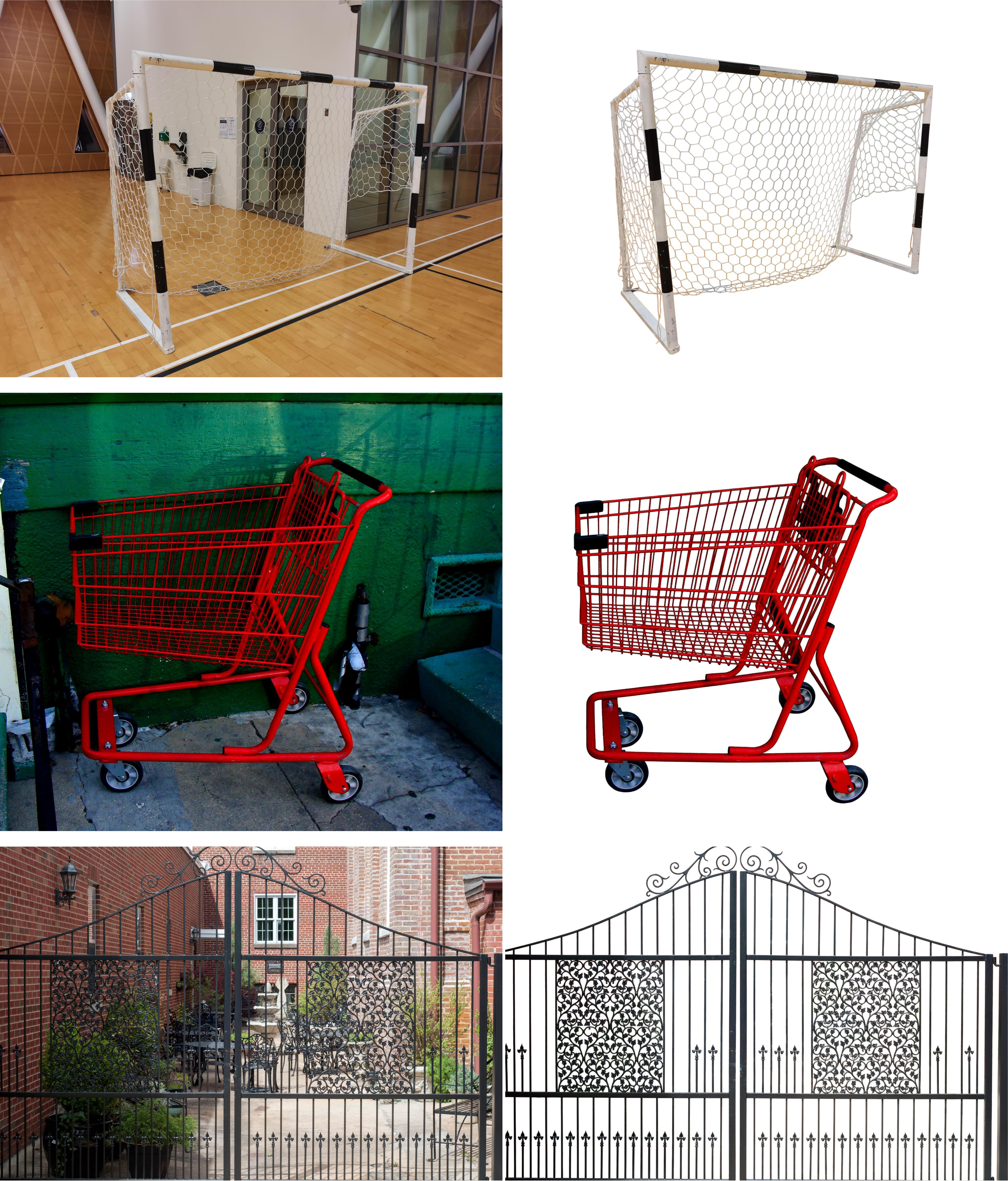}
    \caption{FlowDIS results at $2048 \times 2048$ px resolution on highly detailed samples from DIS-TE4.}
    \label{fig:supp:high_res_samples}
\end{figure*}

We increase the inference resolution of FlowDIS beyond $1024 \times 1024$~px on DIS-TE4, the most challenging subset of DIS5K \cite{qin2022highly}, which contains numerous objects with highly detailed structures. As shown in \cref{tab:supp:resolution}, although FlowDIS was trained only at $1024 \times 1024$~px, its performance improves consistently with higher-resolution inference. \cref{fig:supp:high_res_samples} shows qualitative results obtained with $2048 \times 2048$ px inference on samples with very high levels of detail.

%%%%%%%%%%%%%%%%%%%%%%%%%%%%%%%%%%%%%%%%%%%%%%%%%%%

\end{document}